\documentclass{article}

\usepackage{microtype}
\usepackage{graphicx}
\usepackage{booktabs} %

\usepackage{hyperref}

\usepackage[accepted]{icml2025}

\usepackage{amsmath}
\usepackage{amssymb}
\usepackage{mathtools}
\usepackage{amsthm}
\usepackage{subfig}

\usepackage{url}            %
\usepackage{amsfonts}       %
\usepackage{nicefrac}       %
\usepackage{microtype}      %
\usepackage{xcolor}         %
\usepackage{wrapfig}

\usepackage{enumitem}
\usepackage{diagbox}
\usepackage{bm}
\usepackage{dsfont}
\usepackage{multirow}
\usepackage{varwidth}
\usepackage{pifont}
\usepackage{makecell}
\usepackage{comment}
\usepackage{color}
\usepackage[colaction]{multicol}
\usepackage{colortbl}
\usepackage{algorithm}%
\usepackage{xspace}
\usepackage{rotating}

\usepackage{adjustbox}
\usepackage{threeparttable}
\usepackage{listings}

\usepackage[capitalize,noabbrev]{cleveref}

\let\classAND\AND
\let\AND\relax
\let\classOR\OR
\let\OR\relax
\usepackage{algorithmic}

\let\AND\classAND

\let\OR\classOR
\AtBeginEnvironment{algorithmic}{\let\AND\algoAND}
\AtBeginEnvironment{algorithmic}{\let\OR\algoOR}

\floatname{algorithm}{Algorithm}
\renewcommand{\algorithmicrequire}{\textbf{Input:}}
\renewcommand{\algorithmicensure}{\textbf{Output:}}
\renewcommand{\algorithmicendif}{\textbf{end}}
\definecolor{Gray}{gray}{0.93}

\definecolor{gc_pink}{HTML}{FF6F79}
\definecolor{gc_blue}{HTML}{6FB0FF}
\definecolor{gc_gray}{HTML}{D9D9D9}
\definecolor{gc_dark_pink}{HTML}{99262E}
\definecolor{gc_dark_blue}{HTML}{265A99}
\definecolor{gc_dark_gray}{HTML}{999999}
\definecolor{comment_color}{HTML}{1B8F44}
\definecolor{comment_color_2}{RGB}{64,128,128}
\definecolor{personal_blue}{RGB}{237,245,255}

\newcommand{\LineComment}[1]{\vspace*{0.5em}\small\textcolor{comment_color_2}{\textit{\# #1}}}

\definecolor{codegreen}{rgb}{0,0.6,0}
\definecolor{codegray}{rgb}{0.5,0.5,0.5}
\definecolor{codepurple}{rgb}{0.58,0,0.82}
\definecolor{backcolour}{rgb}{0.95,0.95,0.92}
\definecolor{brickred}{rgb}{0.8, 0.25, 0.33}
\definecolor{linkColor}{rgb}{0,0.1254902,0.37647059}

\lstdefinestyle{mystyle}{
    backgroundcolor=\color{backcolour},   
    commentstyle=\color{codegreen},
    keywordstyle=\color{magenta},
    numberstyle=\tiny\color{codegray},
    stringstyle=\color{codepurple},
    basicstyle=\ttfamily\footnotesize,
    breakatwhitespace=false,         
    breaklines=true,                 
    captionpos=false,                    
    keepspaces=true,                 
    numbersep=5pt,                  
    showspaces=false,                
    showstringspaces=false,
    showtabs=false,                  
    tabsize=2
}

\usepackage{caption}
\usepackage{minitoc}

\newcommand{\wh}{\widehat}

\renewcommand{\hat}{\wh}

\DeclareMathOperator*{\AND}{\mathrm{AND}}
\DeclareMathOperator*{\OR}{\mathrm{OR}}

\newcommand{\diff}[1]{\textcolor{blue}{#1}}

\newcommand{\method}{MMInference}
\newcommand{\methodall}{MMInference}
\icmltitlerunning{MMInference: Accelerating Pre-filling for Long-Context VLMs via Modality-Aware Permutation Sparse Attention}

\begin{document}

\twocolumn[
\icmltitle{\methodall{}: Accelerating Pre-filling for Long-Context \\ Visual Language Models via Modality-Aware Permutation Sparse Attention}

\icmlsetsymbol{equal}{*}
\icmlsetsymbol{cor}{$\mathsection$}

\begin{icmlauthorlist}
\icmlauthor{Yucheng Li}{uos,equal}
\icmlauthor{Huiqiang Jiang}{ms,cor}
\icmlauthor{Chengruidong Zhang}{ms}
\icmlauthor{Qianhui Wu}{ms}
\icmlauthor{Xufang Luo}{ms}
\icmlauthor{Surin Ahn}{ms}
\icmlauthor{Amir H. Abdi}{ms}
\icmlauthor{Dongsheng Li}{ms}
\icmlauthor{Jianfeng Gao}{ms}
\icmlauthor{Yuqing Yang}{ms}
\icmlauthor{Lili Qiu}{ms}
\end{icmlauthorlist}

\icmlaffiliation{ms}{Microsoft Corporation}
\icmlaffiliation{uos}{University of Surrey}

\icmlcorrespondingauthor{Huiqiang Jiang}{hjiang@microsoft.com}

\icmlkeywords{Machine Learning, ICML}

\vskip 0.3in
]

\printAffiliationsAndNotice{\icmlEqualContribution} %

\begin{abstract}
The integration of long-context capabilities with visual understanding unlocks unprecedented potential for Vision Language Models (VLMs). However, the quadratic attention complexity during the pre-filling phase remains a significant obstacle to real-world deployment. To overcome this limitation, we introduce \textbf{\method{}} (\textit{Multi-modality Million tokens Inference}), a dynamic sparse attention method that accelerates the pre-filling stage for long-context multi-modal inputs. First, our analysis reveals that the temporal and spatial locality of video input leads to a unique sparse pattern, the \textit{Grid pattern}.
Simultaneously, VLMs exhibit markedly different sparse distributions across different modalities. We introduce a permutation-based method to leverage the unique Grid pattern and handle modality boundary issues. By offline search the optimal sparse patterns for each head, \method{} constructs the sparse distribution dynamically based on the input. We also provide optimized GPU kernels for efficient sparse computations. Notably, \method{} integrates seamlessly into existing VLM pipelines without any model modifications or fine-tuning. Experiments on multi-modal benchmarks—including Video QA, Captioning, Vision-NIAH, and Mixed-Modality-NIAH—with state-of-the-art long-context VLMs (LongVila, Llava-Video, VideoChat-Flash, Qwen2.5-VL) show that \method{} accelerates the pre-filling stage by up to $8.3\times$ at 1M tokens while maintaining accuracy.
Our code is available at \url{https://aka.ms/MMInference}.
\end{abstract}

\begin{figure}[htb]
  \vspace{-10pt}
  \centering
    \includegraphics[height=0.9\columnwidth]{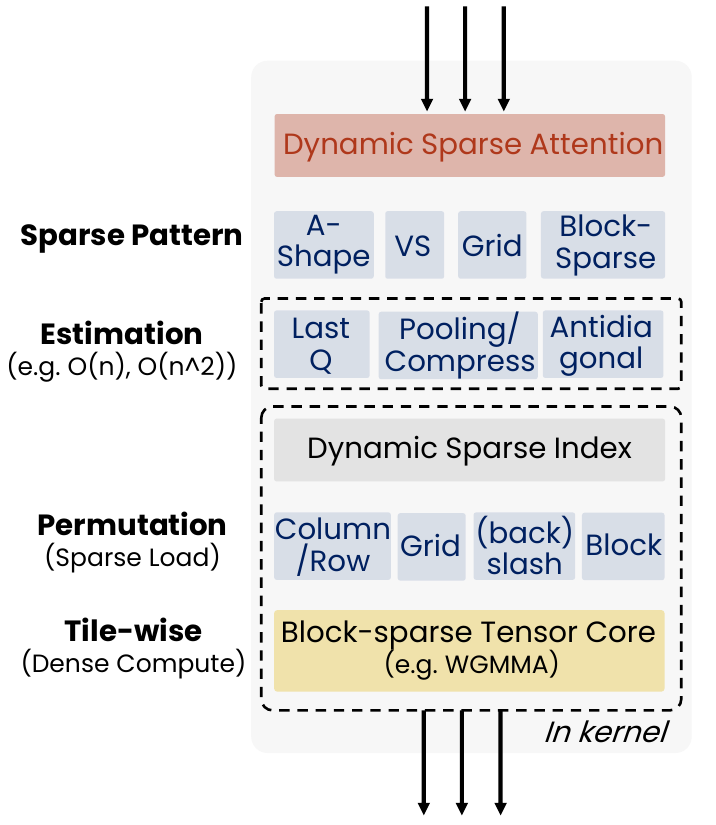}
  \caption{Dynamic sparse attention pipelines leverage sparse loading with dense computation~\cite{zheng2023pit} to enable hardware-efficient acceleration. \methodall{} adopts a bottom-up system-algorithm co-design that accounting for both the mathematical equivalence constraints of sparse loading and the locality properties of real-world attention patterns.
  }
  \label{fig:dynamic_sparse_pipelines}
  \vspace{-10pt}
\end{figure}

\section{Introduction}

Scaling the context size of Vision Language Models (VLMs) allows them to handle extended temporal information from long video and text inputs, which is crucial for various applications including robotics \cite{black2024,prasad2024,cheang2024}, autonomous driving \cite{hu2023,wang2023,gao2024}, and healthcare \cite{liu2024}. In addition, \citet{zhang2024} and \citet{xue2024longvila} show that scaling the context size of VLMs can improve the resolution in the temporal dimension and lead to better performance in video understanding tasks.

\begin{figure*}[htb]
  \vspace{-10pt}
  \centering
  \subfloat[VLMs' attention incurs heavy cost.]{
    \label{sfig:latency}
    \includegraphics[height=0.53\columnwidth]{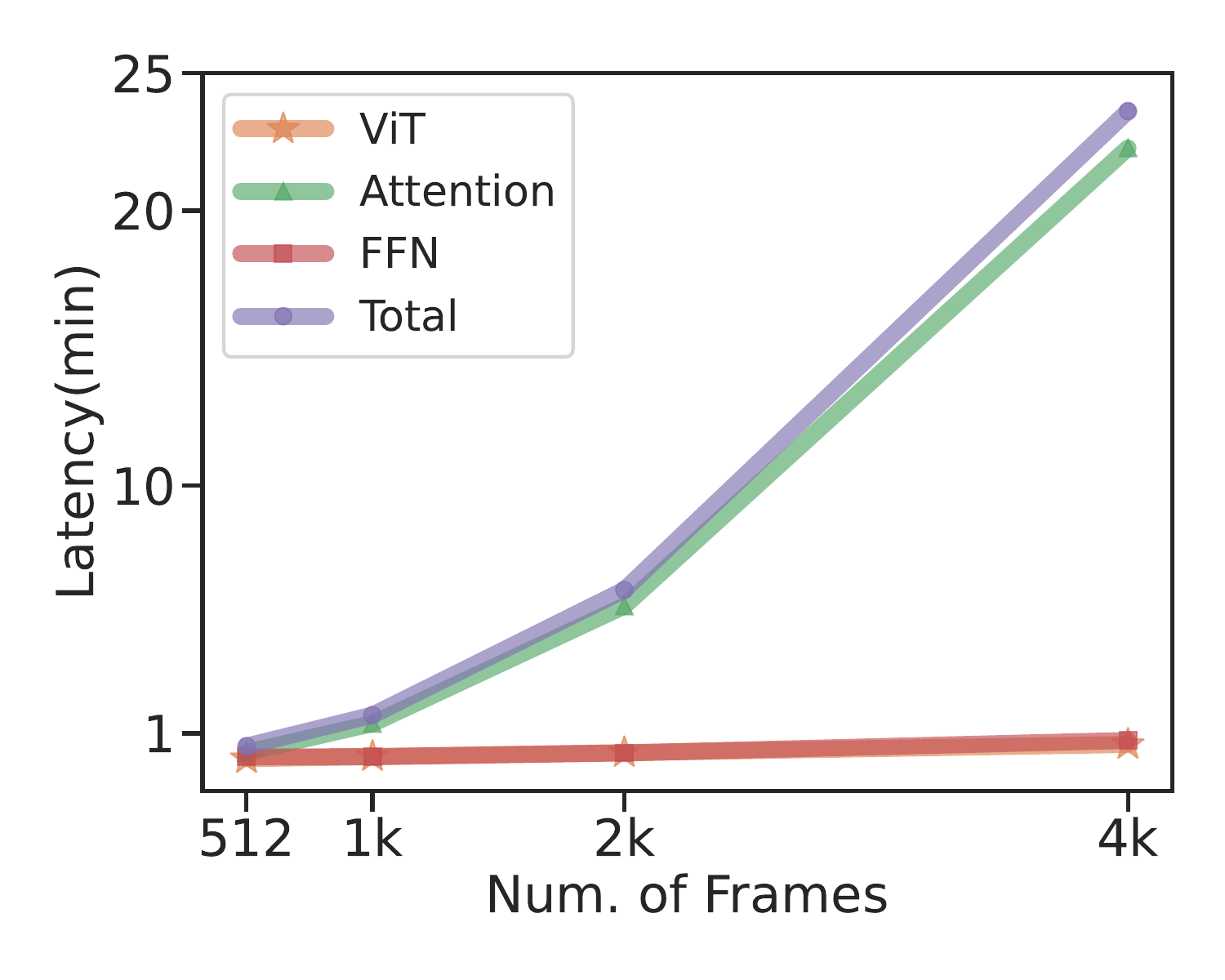}}
    \hspace{0.05em}
  \subfloat[VLMs' attention is sparse.]{
    \label{sfig:sparsity_vlm}
    \includegraphics[height=0.52\columnwidth]{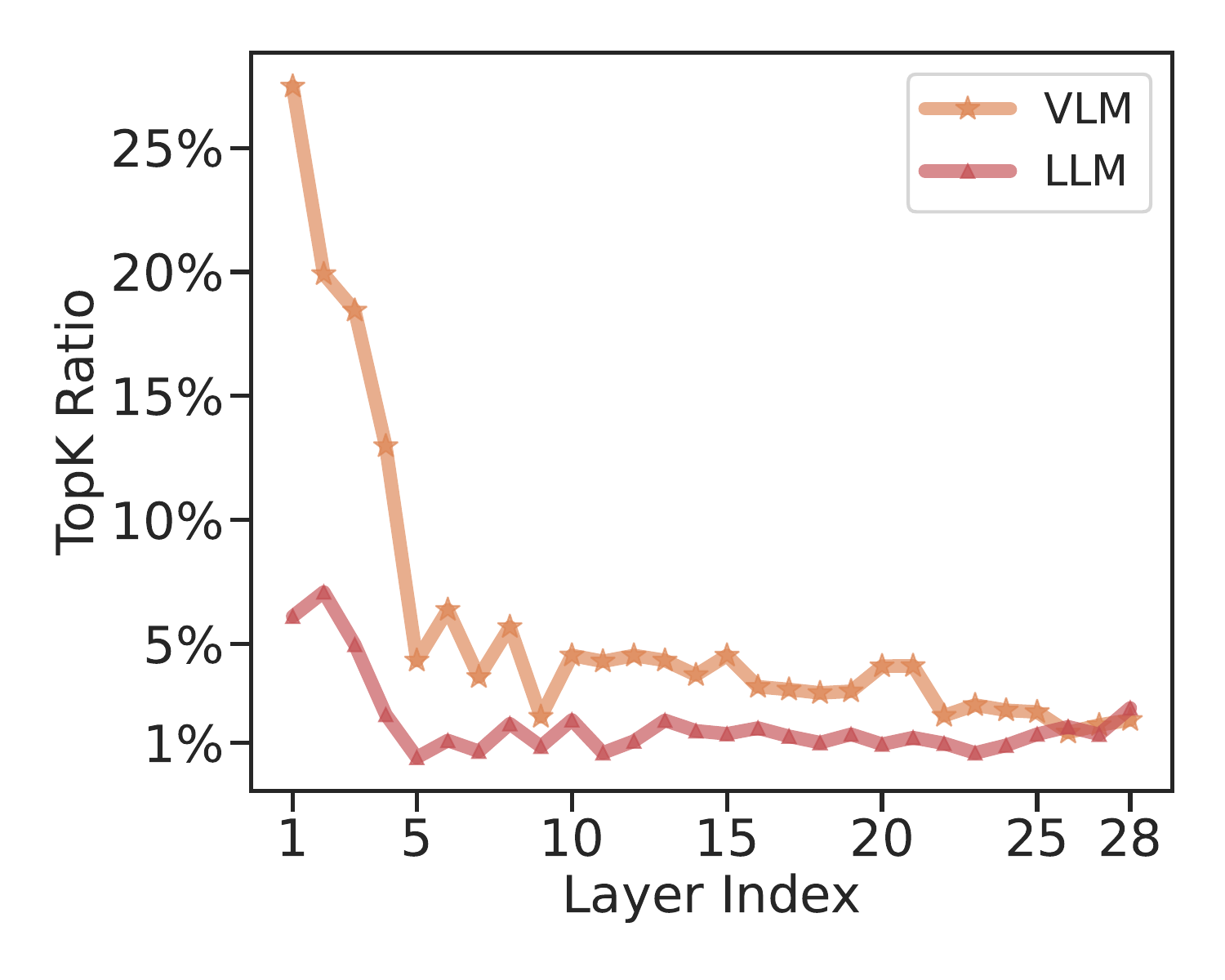}}
  \subfloat[Sparsity of VLMs' attention is dynamic.]{
    \label{sfig:dynamic_sparsity}
    \includegraphics[height=0.54\columnwidth]{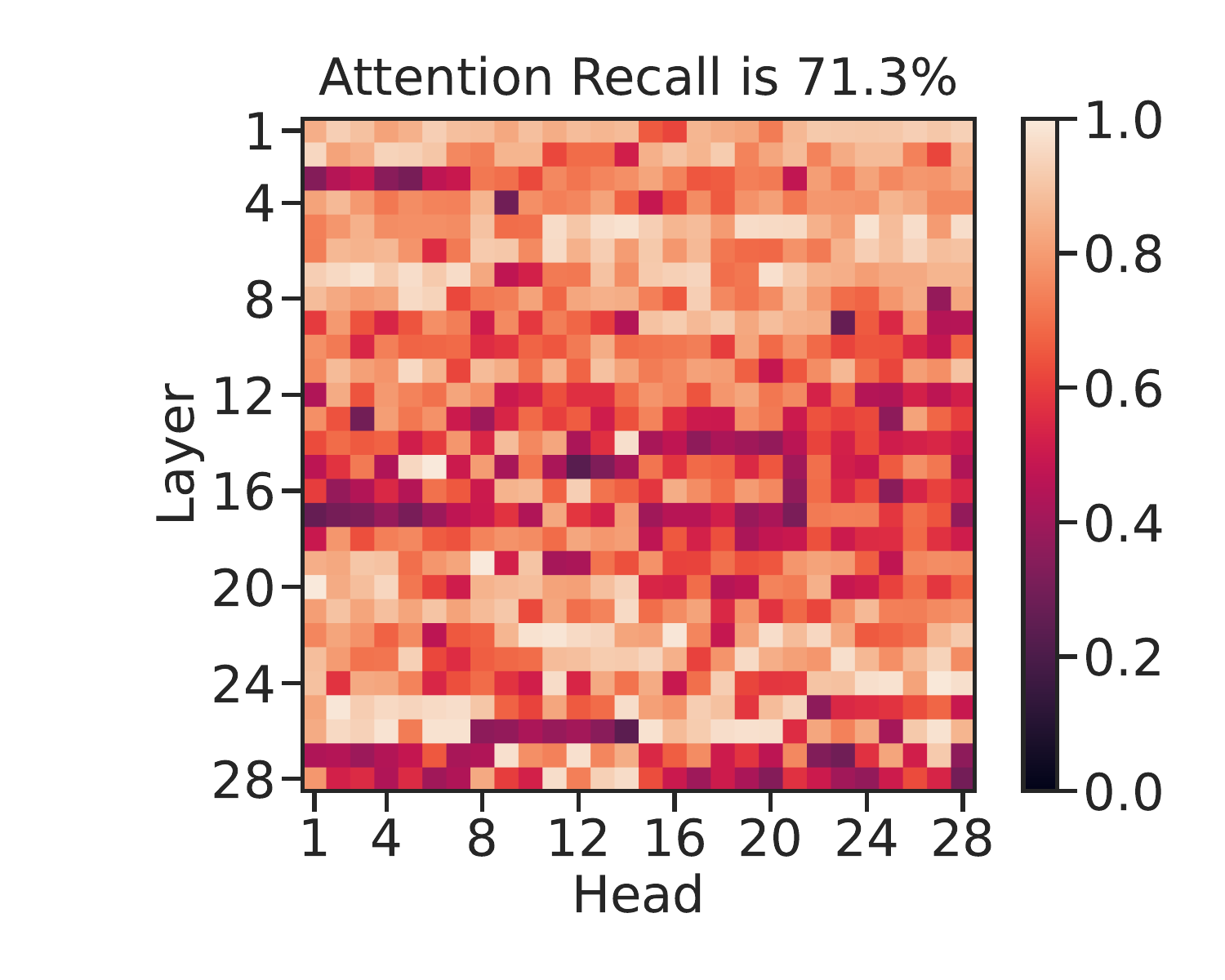}}
  \caption{(a) Latency breakdown of the pre-filling stage, with 256 tokens per frame. 
  (b) How much element in attention needs to be computed to achieve 95\% recall in a 128k context.
  (c) Low attention recall when reusing the top-k indices from a different request.
  Visualizations are based on LongVILA-7B-1M~\cite{xue2024longvila} with a single A100.
  }
  \label{fig:motivations_sparsity}
  \vspace{-10pt}
\end{figure*}

However, due to the quadratic complexity of attention, processing long multi-modal inputs (i.e., the pre-fill stage) can take minutes prior to auto-regressive decoding. As shown in Fig.~\ref{sfig:latency}, this leads to significant Time-to-First-Token latency, which hinders the wide adoption of long-context VLMs in real-world applications. Previous work \cite{child2019,liu2022,liu2024retrievalattention,yuan2025native,lu2025moba} reveals that attention matrices are typically sparse, prompting the development of sparse attention methods such as Sparse Transformer \cite{child2019}, Swin Transformer \cite{liu2021}, and StreamingLLM~\cite{xiao2024efficient}.
More recently, MInference~\cite{jiang2024minference} proposes to use dynamic sparse attention that estimates the sparse index online, and leverages optimized GPU kernels for end-to-end acceleration. However, these methods fail to exploit the unique sparse patterns in long-context VLMs, and struggle with mixed or interleaved modalities, limiting their applicability without compromising performance.

\begin{figure*}[htb]
\vspace{-5pt}
  \centering
  \subfloat[Grid pattern.]{
    \label{sfig:grid}
    \includegraphics[height=0.68\columnwidth]{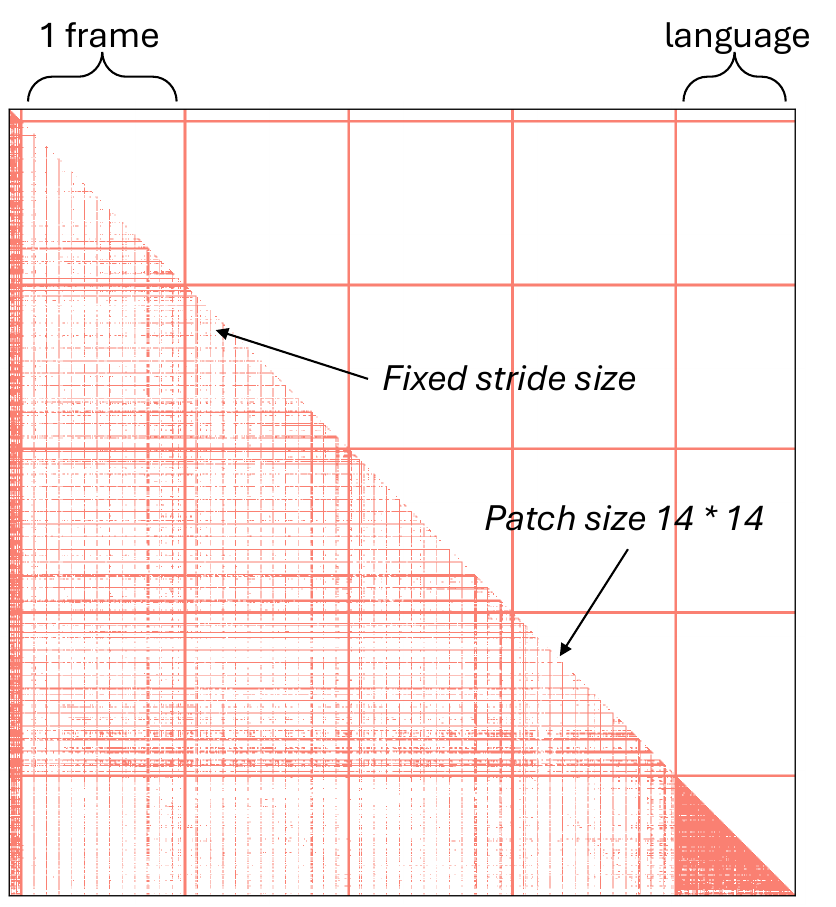}}
    \hspace{0.21em}
  \subfloat[Q-Boundary pattern.]{
    \label{sfig:q_boundary}
    \includegraphics[height=0.68\columnwidth]{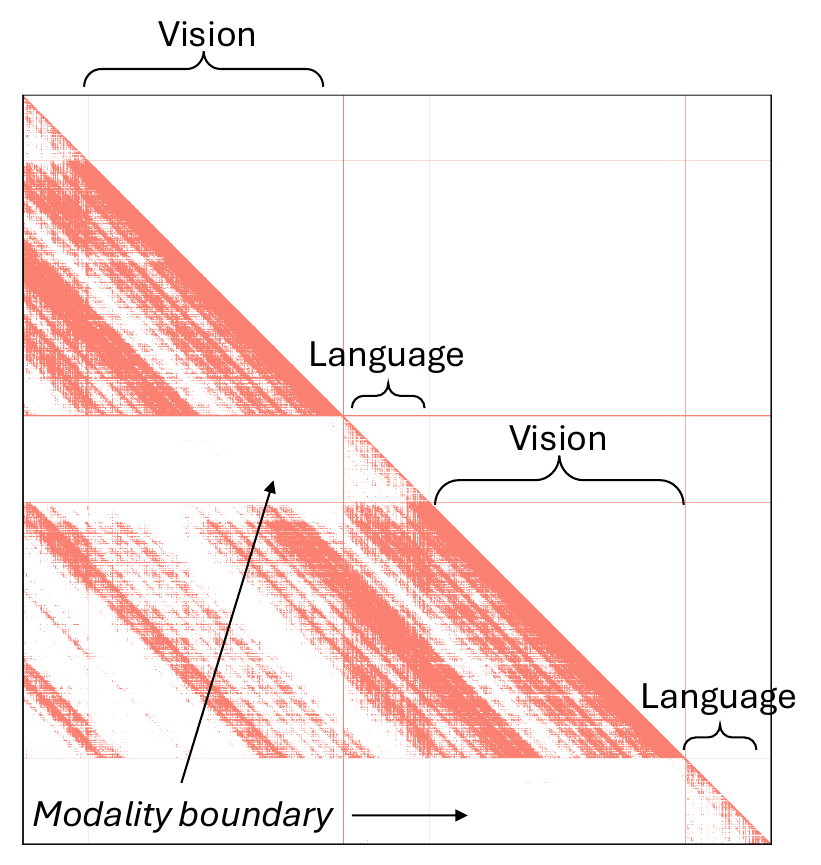}}
  \subfloat[2D-Boundary pattern.]{
    \label{sfig:2d_boundary}
    \includegraphics[height=0.65\columnwidth]{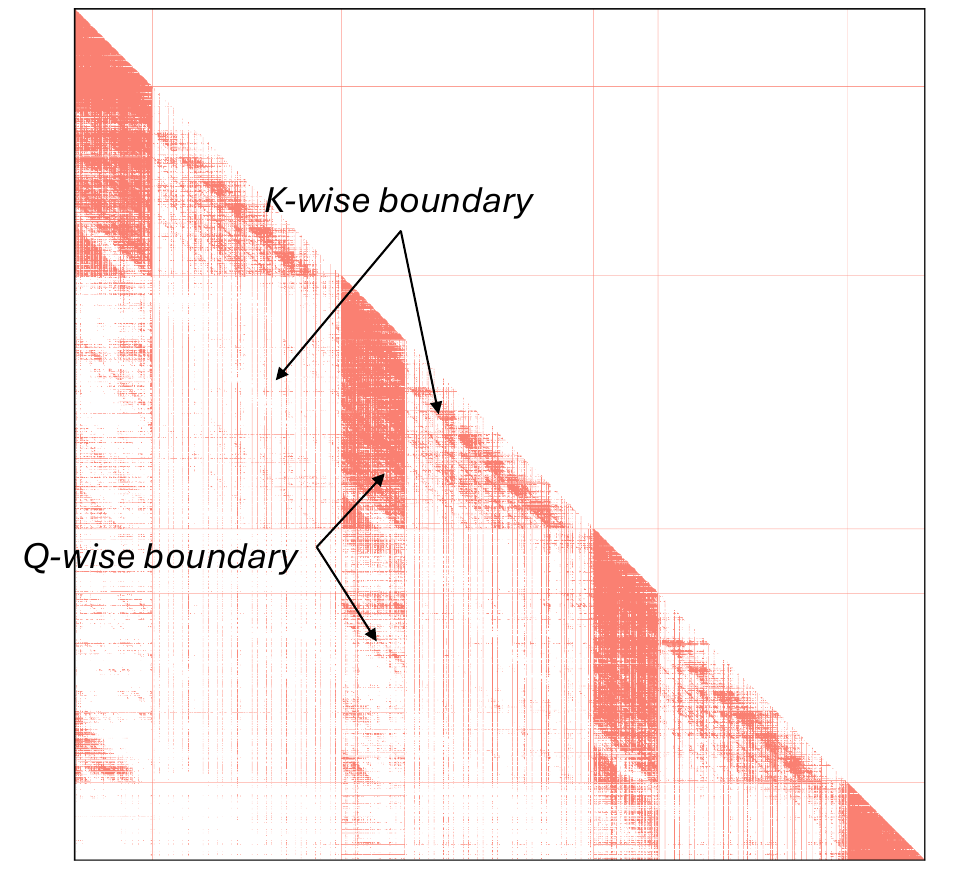}}\\
  \subfloat[Permuted Grid pattern.]{
    \label{sfig:grid_permutation}
    \includegraphics[height=0.65\columnwidth]{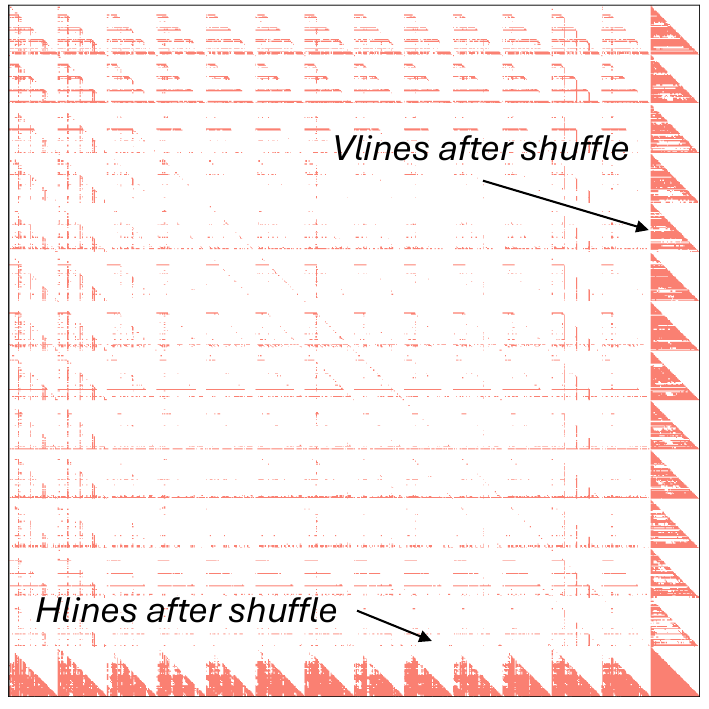}}
    \hspace{0.21em}
  \subfloat[Permuted Q-Boundary pattern.]{
    \label{sfig:q_boundary_permutation}
    \includegraphics[height=0.65\columnwidth]{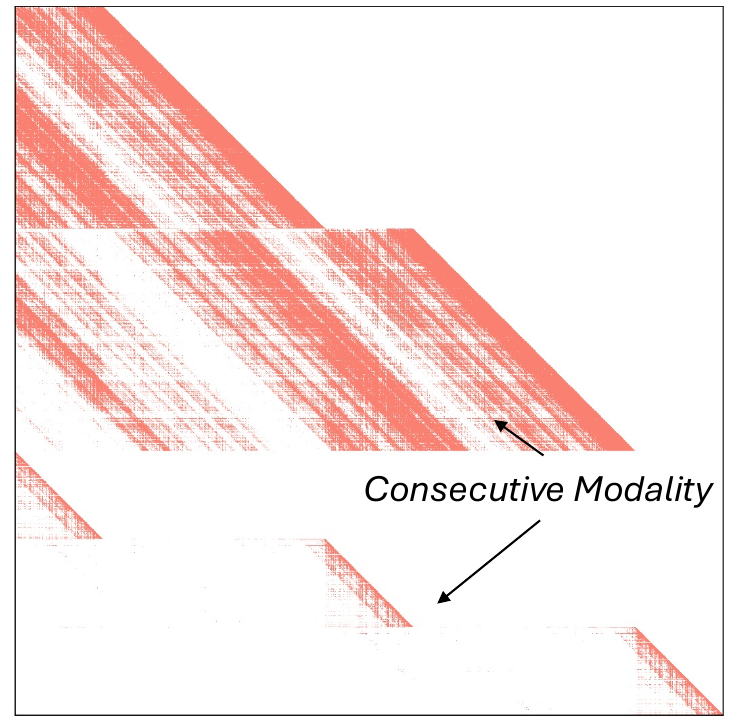}}
    \hspace{0.21em}
  \subfloat[Permuted 2D-Boundary pattern.]{
    \label{sfig:2d_boundary_permutation}
    \includegraphics[height=0.65\columnwidth]{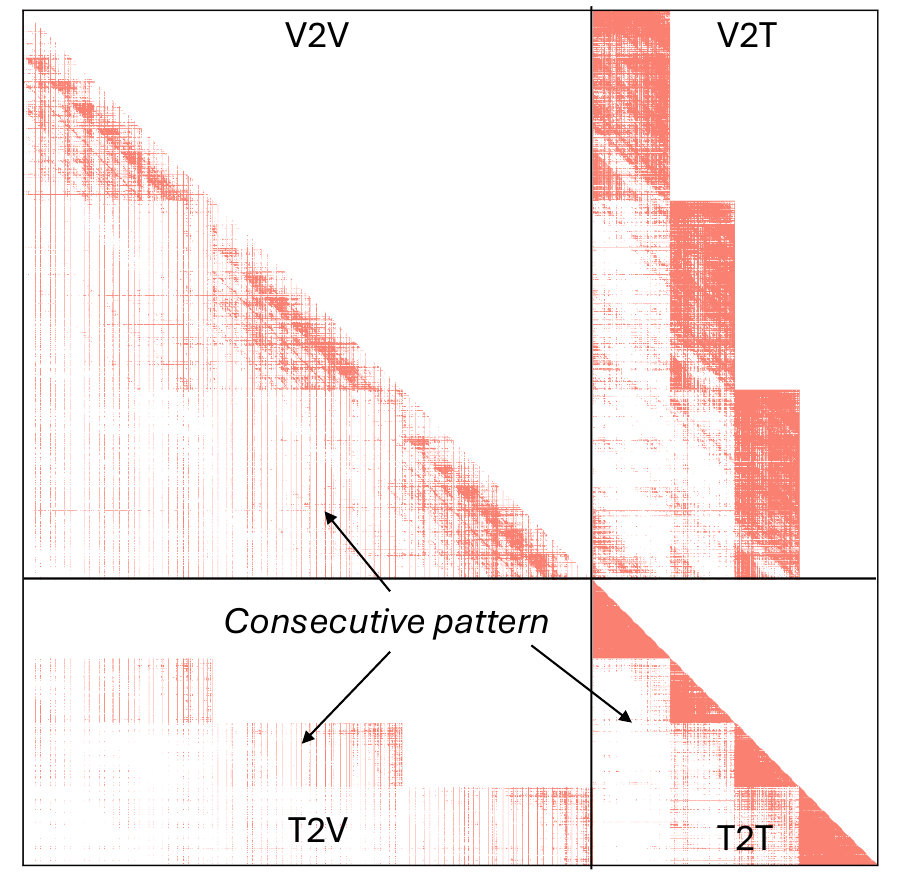}}
  \caption{Visualization of pre- vs. post-permutation sparsity attention patterns in VLMs. 
  }
  \label{fig:motivations_pattern}
  \vspace{-10pt}
\end{figure*}

Unlike long-text contexts, video and image inputs in VLMs exhibit spatiotemporal locality, forming grid-like attention patterns with evenly spaced vertical and horizontal lines (Fig.~\ref{sfig:grid}). In mixed-modality inputs, clear modality boundaries emerge: attention across modalities diverges significantly from intra-modality attention (Fig.~\ref{sfig:q_boundary}). These factors pose unique challenges for exploiting sparsity to accelerate the pre-fill stage.

In this paper, we present \methodall{}, a permutation-based dynamic sparse attention method that significantly reduces attention FLOPs, accelerating the pre-fill stage of long-context VLMs. First, \methodall{} identifies the grid heads and leverages a \textit{row- and column-wise permutation} to gather the sparse grid for efficient hardware computation. Next, we detect Query-boundary and 2D-boundary patterns to address inter-modality boundary issues, and apply a \textit{modality-wise permutation} to isolate intra-modality regions. This results in a consecutive sparse index within each modality, permitting efficient hardware implementation of sparse computing. Finally, a \textit{Modality-Aware Sparse Attention Search Algorithm} is devised to fine-tune both inter- and intra-modality patterns offline, to optimize performance with minimal overhead.

We conduct extensive experiments using four state-of-the-art long-context VLMs, Llava-Video \cite{zhang2024}, LongVila \cite{xue2024longvila}, VideoChat-Flash~\cite{li2024videochat} and Qwen2.5-VL~\cite{bai2025qwen2}, across diverse video understanding tasks such as video captioning \cite{maaz2024}, video question answering \cite{yu2019,xiao2021,mangalam2023,fu2024}, and video information retrieval \cite{zhang2024long}. Additionally, we propose the Mixed-Modality Needle in a Hackathon task to assess multi-modal input performance. Our method effectively addresses modality boundaries, significantly accelerates the prefilling stage, and maintains high accuracy. With a 1M-length context, it achieves speedups of up to 8.3× and 1.7× over FlashAttention-2 and MInference, respectively.

\section{Attention Heads in VLMs}
\label{sec:motivation}

The sparsity of the attention operation in pre-trained text-only LLMs, particularly in long-context scenarios, has been extensively studied~\cite{wu2024retrieval,ribar2024sparq,jiang2024minference,li2024snapkv}, showing that only 3\% of attention weights are activated while achieving a recall rate of 96.8\%. Similarly, VLMs also demonstrate notable dynamic sparsity in long-context scenarios. This section examines the shared and distinct properties of text-only and multi-modal LLMs in long-context scenarios, focusing on attention sparsity, sparse patterns, and modality boundaries.

\subsection{Multi-modality Attention is Dynamically Sparse}

As illustrated in Fig.~\ref{sfig:latency}, for a 128k × 128k attention matrix in VLMs, retaining only the top 5.78\% of attention weights on average suffices to recall 95\% of total attention, indicating that each token attends only to a limited subset of tokens, even in long sequences. However, VLMs exhibit lower sparsity than text-only LLMs, where only 1.79\% of weights achieve a 95\% recall rate. Notably, the bottom layers in VLMs (e.g., the first four layers in LongVila) show reduced sparsity. Yet, due to variability across attention heads, 52.3\% of heads in VLMs require less than 2\% of attention to be recalled. This highlights substantial computational redundancy in VLMs, especially in long-context scenarios.

Similarly to LLMs, while the sparse nature of attention matrices remains consistent across inputs, the specific distributions of sparse attention are highly dynamic. As shown in Fig.~\ref{sfig:dynamic_sparsity}, reusing top-k indices for 95\% attention recall (derived from Fig.~\ref{sfig:sparsity_vlm}) across different contexts leads to a significant drop in performance.

\subsection{The Grid Head in VLMs}

In long-context language modeling, efficient attention mechanisms like sliding window attention~\cite{jiang2023mistral} and StreamingLLM~\cite{xiao2024efficient} exploit the locality property of text sequences. However, multi-modal inputs introduce unique geometric structures that redefine locality. As shown in~\citet{child2019}, image patches exhibit locality along both vertical and horizontal directions, forming local window and slash-like patterns. Similarly, video inputs maintain locality across temporal and spatial dimensions, with frame-based sampling yielding more regular and predictable patterns.

We observe that certain VLM attention heads exhibit a \textbf{grid pattern}. While the grid's stride and starting position vary with context, the horizontal and vertical lines are evenly spaced and often symmetrical—a distinct behavior compared to text-only LLMs~\cite{jiang2024minference,lai2025flexprefill}. Fig.~\ref{sfig:grid} visualizes a grid head, demonstrating how local tokens in temporal and spatial dimensions are evenly distributed within the attention map, with attention focused primarily on these local tokens.

\begin{figure*}[htb]
    \centering
    \vspace{-5pt}
    \resizebox{1.6\columnwidth}{!}{
    \includegraphics[width=\linewidth]{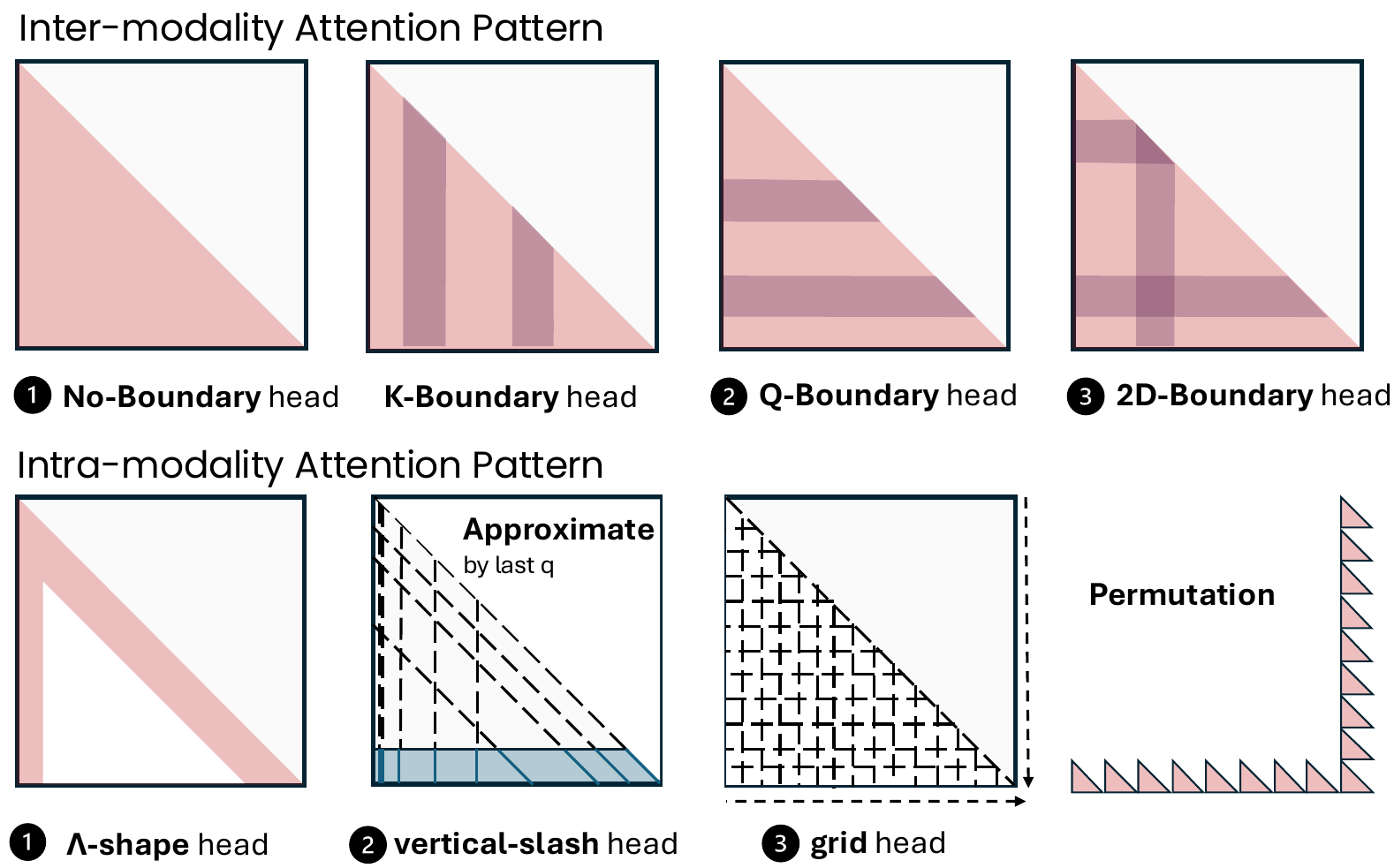}
    }
    \caption{The framework of \methodall{}, encompassing both inter- and intra-modality sparse attention patterns.}
    \label{fig:framework}
    \vspace{-5pt}
\end{figure*}

\subsection{Modality Boundaries in Multi-Modal Input}

The input format of VLMs differs significantly from text-only LLMs. A dedicated vision encoder generates visual representations, which are processed alongside text embeddings by the LLM. Despite pretraining on large-scale datasets, the interactions and processing patterns between modalities vary considerably, leading to distinct modality boundaries in attention~\cite{tu2024vlcache}, as illustrated in Fig.~\ref{sfig:q_boundary} and \ref{sfig:2d_boundary}.

Specifically, we observe two key characteristics:  
1) \textit{Intra-modality consistency}: Attention within each modality follows a consistent pattern. For instance, the vision region in Fig.~\ref{sfig:q_boundary} exhibits a clear slash pattern, where critical elements are effectively clustered.  
2) \textit{Modality-separated continuity}: Patterns within a modality can be interrupted by boundaries from other modalities. As shown in Fig.~\ref{sfig:q_boundary}, vision slashes are segmented by the boundary introduced by the language region.

We categorize the modality boundary patterns of VLMs into four distinct types: {No-Boundary}, {K-Boundary}, {Q-Boundary}, and {2D-Boundary}, as illustrated in Figs.~\ref{fig:motivations_pattern} and \ref{fig:framework}.  
1) \textbf{No Boundary} and \textbf{K-Boundary} exhibit either no clear modality boundary or a boundary only along the key dimension, as shown in Fig.~\ref{fig:add_boundary_pattern}. Since continuity is maintained along the query dimension, these heads can be efficiently handled using intra-modality sparse patterns.  
2) \textbf{Q-Boundary} refers to attention modality boundaries across the query dimension. For example, in Fig.~\ref{sfig:q_boundary}, sparse patterns like Text-to-Video and Video-to-Video appear interconnected, forming a trapezoidal structure, while a clear boundary separates Visual-to-Text and Text-to-Visual attention.  
3) \textbf{2D-Boundary} occurs when modality boundaries are present in both query and key dimensions. As shown in Fig.~\ref{sfig:2d_boundary}, the 2D modality boundary segments attention weights into distinct blocks.  
Additionally, our analysis of Audio LMs~\cite{chu2024qwen2} and end-to-end multimodal LMs~\cite{xu2025qwen2,li2025baichuanomni15technicalreport} reveals that the cross-modality boundary phenomenon persists across these architectures. These boundaries pose unique challenges and hinder direct application of existing sparse attention methods to multi-modal inputs.

\subsection{Sparse Distributions Continuity Across Boundaries}
\label{subsec:index_continue}

Although sparsity patterns in VLMs are often discontinuous across modalities due to modality boundaries, we find that sparsity distributions can remain continuous across these boundaries and extrapolate to other regions of the same modality. For example, in Fig.~\ref{sfig:q_boundary}, 
the slash lines maintain the same relative position across different areas of the vision modality. In a more complex case, Fig.~\ref{sfig:2d_boundary} shows interleaved vision and text modalities forming a mixed structure. However, by spatially aggregating regions of the same modality, we observe that sparsity patterns can extend beyond local regions and often exhibit global extrapolation potential. The upper-left region in Fig.~\ref{sfig:2d_boundary} exemplifies this, where the grid pattern, initially separated by textual boundaries, becomes consecutive after spatial clustering in both row and column dimensions. To validate this observation, we conducted a quantitative attention recall experiment on mixed-modality inputs, as detailed in \S\ref{subsec:analysis}.

\section{\methodall{}}
\label{sec:method}

Following the analysis in \S\ref{sec:motivation}, we propose \methodall{} to accelerate the pre-filling stage of long-context VLMs as shown in Fig.~\ref{fig:framework}. The framework consists of three modules, covering both inter- and intra-modality sparse patterns: 1) the novel Grid sparse attention, together with the A-shape and Vertical-Slash patterns \cite{jiang2024minference} forms the intra-modality attention; 2) Q-Boundary and 2D-Boundary mix-modality patterns; 3) Modality-aware sparse attention search algorithm. We first perform offline pattern search to identify different patterns for each attention head. Then we use online dynamic sparse approximation to build the sparse index, and finally we perform dynamic sparse computation using optimized GPU kernels.

\subsection{Grid Head in Multi-Modality}

To better leverage the inductive bias in visual modalities (e.g., images, videos) and the vertical and horizontal structures in attention patterns, we propose a permutation-based dynamic sparse attention for grid head, as shown in Algo.~\ref{alg:grid_attention}.

\restylefloat{algorithm}

\begin{minipage}{0.42\textwidth}
\begin{algorithm}[H]
\captionsetup[algorithm]{singlelinecheck=off}
\caption{Grid Head}
\label{alg:grid_attention}
\begin{algorithmic}
  \STATE {\bfseries Input:} $\boldsymbol{Q},\boldsymbol{K},\boldsymbol{V} \in \mathbb{R}^{S \times d_h}$, stride space $s_g\in \phi_g$

    \LineComment{Approximate stride and phase (last\_q = 64)}
    \STATE $\boldsymbol{\hat{A}} \gets \mathrm{softmax}\left(\boldsymbol{Q}_{[-\text{last\_q}:]} \bm{K}^{\top} / \sqrt{d} + \bm{m}_{\text{casual}} \right)$
    
    \LineComment{Online search grid stride and phase}
    \STATE $\boldsymbol{b}_r, \gets 0$
    \FOR{$i \gets 1$ to $|\phi_g|$}
        \IF{$\mathrm{max} (\mathrm{view}(\boldsymbol{\hat{A}}, s_{g,i})) >\boldsymbol{b}_r$}
        \STATE $s_g \gets s_{g,i}, p_g \gets \mathrm{argmax}(\mathrm{view}(\boldsymbol{\hat{A}}, s_{g,i}))$
        \STATE $\boldsymbol{b}_r \gets \mathrm{max} (\mathrm{view}(\boldsymbol{\hat{A}}, s_{g,i}))$
        \ENDIF
    \ENDFOR

    \LineComment{Permute Q, K, V tensors}
    \STATE $\boldsymbol{\overline{Q}}, \boldsymbol{\overline{K}}, \boldsymbol{\overline{V}} \gets \mathrm{permute}\left(\boldsymbol{Q}\right),\mathrm{permute}\left(\boldsymbol{K}\right), \mathrm{permute}\left(\boldsymbol{V}\right)$
 
    \LineComment{Dynamic block sparse attention w/ FlashAttention (only the last and rightmost block)}
    \STATE $\boldsymbol{A} \gets \mathrm{softmax}\left(\mathrm{sparse}(\boldsymbol{\overline{Q}} \boldsymbol{\overline{K}}^\top, s_g, p_g) / \sqrt{d}\right)$
           
    \LineComment{Sparse mixed scores and values}
    \STATE $\boldsymbol{y} \gets \mathrm{sparse}(\boldsymbol{A} \bm{\overline{V}},  s_g, p_g)$\\
    \STATE $\mathrm{return}\,\,\,\boldsymbol{y}$
   
\end{algorithmic}
\end{algorithm}
\end{minipage}
\vspace{5pt}

Specifically, we first perform an online search to determine the stride and phase of grid pattern. Since only a view operation is applied to the approximate attention matrix $\boldsymbol{\hat{A}}$, the actual latency overhead remains minimal. Next, we use the identified grid stride and phase to permute the $\boldsymbol{Q}$, $\boldsymbol{K}$, and $\boldsymbol{V}$ tensors to compute sparse attention efficiently (see Fig.~\ref{sfig:grid_permutation}).  
In our implementation, instead of explicitly permuting $\boldsymbol{Q}$, $\boldsymbol{K}$, and $\boldsymbol{V}$, we optimize computational efficiency by dynamically loading and writing these tensors within the kernel, minimizing the overhead associated with tensor transpositions. In addition to Grid sparse attention, we also employ A-shape and Vertical-Slash attention for intra-modality operation, see Appendix \ref{sec:appendix:a_shape_vs} for more details.

\subsection{Hybrid Modality Sparse Attention}

As analyzed in \S\ref{sec:motivation} and illustrated in Fig.~\ref{fig:motivations_pattern}, modality boundaries exist in multi-modal LLMs. We classify these boundaries into four patterns: {No-Boundary}, {K-Boundary}, {Q-Boundary}, and {2D-Boundary}.
As the sparse index is continuous along the query dimension for both the {No-Boundary} and {K-Boundary} heads, we can directly apply the three intra-modality attention globally.
However, for {Q-Boundary} and {2D-Boundary}, \methodall{} uses a permutation-based approach to efficiently handle these modality boundaries.

\paragraph{Q-Boundary Head}
As shown in Fig.\ref{sfig:q_boundary}, Fig.\ref{sfig:q_boundary_permutation}, and \S\ref{subsec:index_continue}, the Q-Boundary pattern shows a clear separation across modality, but the sparse distribution remains continuous within each modality. Building on this insight, we propose a row-wise permutation (Algorithm~\ref{alg:q_boundary}) that groups tokens of the same modality by permuting $\boldsymbol{Q}$, and then applies offline-optimized sparse attention (A-shape, Vertical-Slash, and Grid Head) for intra-modality processing.
Note that we leverage the final segment of \textit{each modality’s} queries to dynamically approximate the sparse indices and extrapolate to the entire modality. This method enables flexibility in handling fragmented multi-modality inputs. Additionally, instead of explicitly permuting tensors, our implementation performs dynamic loading and writing inside the kernel for optimized efficiency.

\restylefloat{algorithm}

\begin{minipage}[htb]{0.42\textwidth}
\begin{algorithm}[H]
\captionsetup[algorithm]{singlelinecheck=off}
\caption{Q-Boundary Head}
\label{alg:q_boundary}
\begin{algorithmic}
  \STATE {\bfseries Input:} $\boldsymbol{Q},\boldsymbol{K},\boldsymbol{V} \in \mathbb{R}^{S \times d_h}$, modality type index $\boldsymbol{i}_m$, modality type set $m\in \phi_m$

    \LineComment{Permute Q tensors based on modality}
    \STATE $\boldsymbol{\overline{Q}} \gets \mathrm{permute}\left(\boldsymbol{Q}, \boldsymbol{i}_m\right)$
 
    \LineComment{Looping over the modalities in query dimension}
    \STATE $\boldsymbol{y} \gets \boldsymbol{0}$
    \FOR{$i \gets 1$ to $|\phi_m|$}
        \STATE
        \LineComment{Intra-modality sparse attention for each modality w/ FlashAttention}
        \STATE $\boldsymbol{A}_{mi} \gets \mathrm{softmax}\left(\mathrm{sparse}(\boldsymbol{\overline{Q}}_{mi} \boldsymbol{K}^\top, \boldsymbol{i}_{mi} ) / \sqrt{d}\right)$
        \STATE $\boldsymbol{y}_{mi} \gets \mathrm{sparse}(\boldsymbol{A}_{mi} \bm{{V}})$\\
        \LineComment{Update the modality output to the final output}
        \STATE $\boldsymbol{y} \gets \boldsymbol{y}_{mi} \cup \boldsymbol{y}$
    \ENDFOR

    \STATE $\mathrm{return}\,\,\,\boldsymbol{y}$
   
\end{algorithmic}
\end{algorithm}
\end{minipage}

\paragraph{2D-Boundary Head}

\begin{table*}[t]
    \centering
    \small
    \caption{Performance (\%) of different models and different methods on video understanding tasks evaluated at frames from 110 to 256.}
    \label{tab:benchmark_results}
    \resizebox{0.99\textwidth}{!}{
    \begin{tabular}{l c c c c c c c c c}
        \toprule
        \multirow{2}{*}{\textbf{Model}} & \multirow{2}{*}{\textbf{FLOPs}} & \textbf{VideoDC} & \textbf{ActNet-QA} & \textbf{EgoSchema} & \textbf{Next-QA} & \textbf{PerceptionTest} & \multicolumn{2}{c}{\textbf{VideoMME}} & \multirow{2}{*}{\textbf{Avg.}} \\
        \cmidrule(lr){3-9}
        & & test & test & test & mc & val & w/o sub. & w/ sub. & \\
        \midrule
        \multicolumn{10}{c}{\textit{Llava-Video-7B \quad \# Frames: 110; Total \# tokens: 20,240}} \\
        \midrule
        \rowcolor{Gray}
        Full Attention & 100\% & 3.66 & 59.6 & 57.0 & 81.2 & 66.1 & 64.7 & 71.0 & 57.6 \\
        SF-fixed & 4.8\% & 3.26 & 57.3 & 53.3 & 79.8 & 62.9 & 59.9 & 67.1 & 54.8\\
        SF-strided & 41.4\% & 3.45 & 58.5 & 56.1 & 80.6 & 64.4 & 61.4 & 68.5 & 56.1 \\
        A-shape & 48.2\% & 3.56 & 56.0 & 51.6 & 79.8 & 65.7 & 54.4 & 65.6 & 53.8\\
        Tri-shape & 49.0\% & 3.58 & 59.3 & 54.5 & 80.3 & 66.1 & 63.6 & 70.1 & {56.7} \\
        VisionZip & 35.2\% & 1.35 & 42.1 & 40.5 & 69.5 & 41.4 & 44.9 & 62.1 & {43.1} \\
        MInference & 78.8\% & 3.64 & 59.6 & 57.0 & 80.6 & 66.1 & 64.6 & 71.0 & {57.5}\\
        \textbf{Ours} & 47.3\% & 3.58 & \textbf{59.8} & \textbf{57.1} & 80.1 & \textbf{66.2} & 64.5 & \textbf{71.8} & \textbf{57.6} \\
        \midrule
        \multicolumn{10}{c}{\textit{LongVILA-7B \quad \# Frames: 256; Total \# tokens: 65,800}} \\
        \midrule
        \rowcolor{Gray}
        Full Attention & 100\% & 2.76 & 59.5 & 61.9 & 80.7 & 58.1 & 60.1 & 65.1 & 55.5 \\
        SF-fixed & 2.2\% & 1.99 & 51.3 & 59.6 & 76.5 & 55.5 & 57.1 & 63.0 & 52.1\\
        SF-strided & 26.6\% & 2.58 & 56.0 & 61.4 & 76.7 & 55.5 & 53.6 & 59.2 & 52.2\\
        A-shape & 29.1\% & 2.75 & 56.6 & 60.9 & 75.0 & 55.3 &  49.1 & 59.6 & 51.3 \\
        Tri-shape & 29.3\% & 2.63 & 58.1 & 62.0 & 77.8 & 56.2 & 59.3 & 63.3 & 54.2\\
        \rowcolor{Gray}
        VisionZip & \multicolumn{8}{c}{OOM} & \\
        MInference & 47.0\% & 2.77 & 59.7 & 62.2 & 79.1 & 57.8 & 60.0 & 65.2 & 55.2\\
        \textbf{Ours} & 31.8\% & \textbf{2.84} & \textbf{60.2}  & 62.2 &\textbf{79.4} & 57.8 & 60.0 & \textbf{65.5} & \textbf{55.4}\\
        \midrule
        \multicolumn{10}{c}{\textit{Qwen2.5-VL-7B-Instruct \quad \# Frames: 256; Total \# tokens: 33,950}} \\
        \midrule
        \rowcolor{Gray}
        Full Attention & 100\% & 3.71 & 58.3 & 64.3 & 85.4 & 68.7 & 64.7 & 71.3 & 59.5 \\
        \textbf{Ours} & 41.3\% & \textbf{3.75} & 58.0  & 63.9 & 84.9 & \textbf{68.9} & \textbf{65.1} & 70.9 & 59.4\\
        \bottomrule
    \end{tabular}
    }
\end{table*}

Beyond Query-Boundary, there are attention heads that exhibit modality boundaries in both query and key dimensions, as shown in Fig.~\ref{sfig:2d_boundary}.
Given a query token, attention to key tokens from different modalities varies significantly, and queries from different modalities focus on keys in highly diverse patterns.
To address 2D modality boundaries, we design a 2D permutation approach that groups $\boldsymbol{Q}$, $\boldsymbol{K}$, and $\boldsymbol{V}$ according to their modalities. This allows us to leverage intra-modality continuity to handle each part of 2D boundary pattern separately and efficiently. We further illustrate this approach in Fig.~\ref{sfig:2d_boundary_permutation} and it detailed in Algorithm~\ref{alg:2d_boundary}. Specifically, we perform permutation on both row- and column-wise for $\boldsymbol{Q}$, $\boldsymbol{K}$, and $\boldsymbol{V}$, and then iteratively traverse each modality pair to compute dynamic
\vspace{-10pt}
\restylefloat{algorithm}

\begin{minipage}[b!]{0.42\textwidth}
\begin{algorithm}[H]
\captionsetup[algorithm]{singlelinecheck=off}
\caption{2D-Boundary Head}
\label{alg:2d_boundary}
\begin{algorithmic}
  \STATE {\bfseries Input:} $\boldsymbol{Q},\boldsymbol{K},\boldsymbol{V} \in \mathbb{R}^{S \times d_h}$, modality type index $\boldsymbol{i}_m$, modality type set $m\in \phi_m$

    \LineComment{Permute Q, K, V tensors based on modality}
    \STATE $\boldsymbol{\overline{Q}} \gets \mathrm{permute}\left(\boldsymbol{Q}, \boldsymbol{i}_m\right), \boldsymbol{\overline{K}} \gets \mathrm{permute}\left(\boldsymbol{K}, \boldsymbol{i}_m\right)$
    \STATE $\boldsymbol{\overline{V}} \gets \mathrm{permute}\left(\boldsymbol{V}, \boldsymbol{i}_m\right)$
 
    \LineComment{Looping over the modalities in pairs}
    \STATE $\boldsymbol{y} \gets \boldsymbol{0}$
    \FOR{$i \gets 1$ to $|\phi_m|$}
        \FOR{$j \gets 1$ to $|\phi_m|$}
        \STATE
        \LineComment{Dynamic sparse attention for each modality pair w/ FlashAttention}
        \STATE $\boldsymbol{m}_{mi,mj} \gets \mathrm{buildmask}(\boldsymbol{i}_{mi}, \boldsymbol{i}_{mj})$
        \STATE $\boldsymbol{A}_{mi,mj} \gets \mathrm{softmax}($
        \STATE $\mathrm{sparse}(\boldsymbol{\overline{Q}}_{mi}\boldsymbol{\overline{K}}_{mj}^\top, \boldsymbol{i}_{mi}, \boldsymbol{i}_{mj}) / \sqrt{d}+ \boldsymbol{m}_{mi,mj})$
        \STATE $\boldsymbol{y}_{mi,mj} \gets \mathrm{sparse}(\boldsymbol{A}_{mi,mj} \bm{\overline{V}}_{mj})$\\
        \LineComment{Update the modality output to the final output}
        \STATE $\boldsymbol{y} \gets \boldsymbol{y}_{mi,mj} \cup \boldsymbol{y}$
        \ENDFOR
    \ENDFOR

    \STATE $\mathrm{return}\,\,\,\boldsymbol{y}$
   
\end{algorithmic}
\end{algorithm}
\end{minipage}
\vspace{5pt}

sparse attention.  
The 2D-Boundary requires constructing an attention mask and searching for sparse patterns in cross-modality regions. For example, in Fig.~\ref{sfig:2d_boundary_permutation}, we build modality boundary indices for Vision-to-Text (bottom-left) and Text-to-Vision (upper-right) attention. This mask index construction is implemented in Triton \cite{tillet2019triton}.

\subsection{Modality-Aware Sparse Attention Search Algorithm}

Due to modality boundaries in VLMs, we propose a modality-aware sparse attention pattern search algorithm (see Algorithm~\ref{alg:search}). The process unfolds in three steps: 1) intra-modality search within each modality following \cite{jiang2024minference}, 2) cross-modality search across all modality pairs, and 3) inter-modality search informed by the results of the first two steps.

\section{Experiments}
\label{sec:experiments}

In this section, we address two key questions:  
(i) \textbf{How effective \methodall{} is?} We evaluate our method on three general long-video tasks: long-video understanding, Video Needle in a Haystack, and Video-Text Needle in a Haystack. These benchmarks cover long-video captioning, open-ended QA, multiple-choice QA, mixed-modality tasks, and retrieval tasks, providing a comprehensive assessment of \methodall{}’s effectiveness across diverse long-video scenarios.  
(ii) \textbf{How efficient \methodall{} is?} We analyze end-to-end latency and its breakdown to thoroughly evaluate the efficiency of \methodall{}.

\subsection{Dataset and Baselines}

\begin{figure*}[htb]
\vspace{-5pt}
  \centering
  \subfloat[\methodall{} in V-NIAH]{
    \label{sfig:needle_our_vniah}
    \includegraphics[width=1\columnwidth]{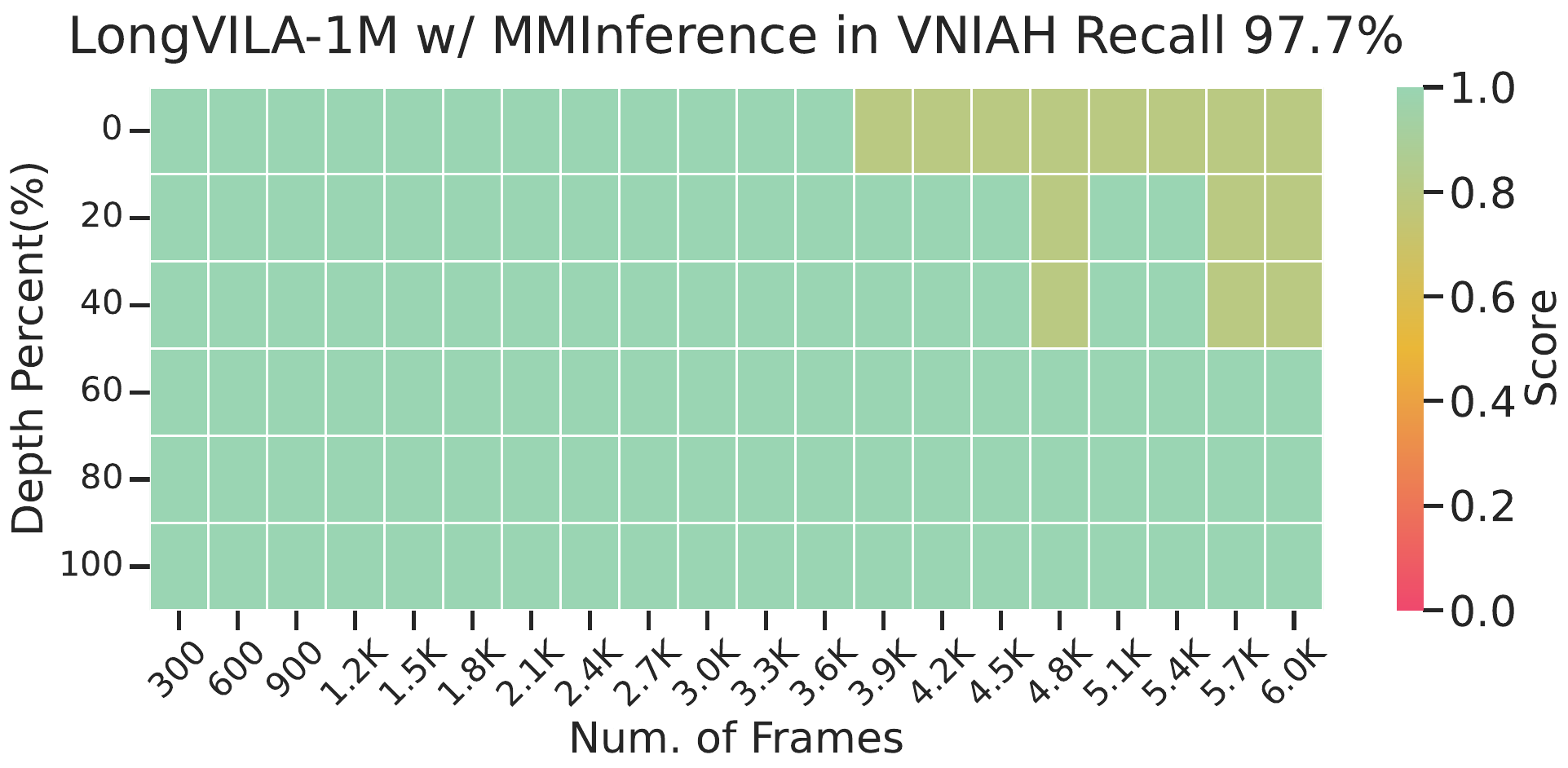}}
  \subfloat[FullAttention in V-NIAH]{
    \label{sfig:needle_dense_vniah}
    \includegraphics[width=1\columnwidth]{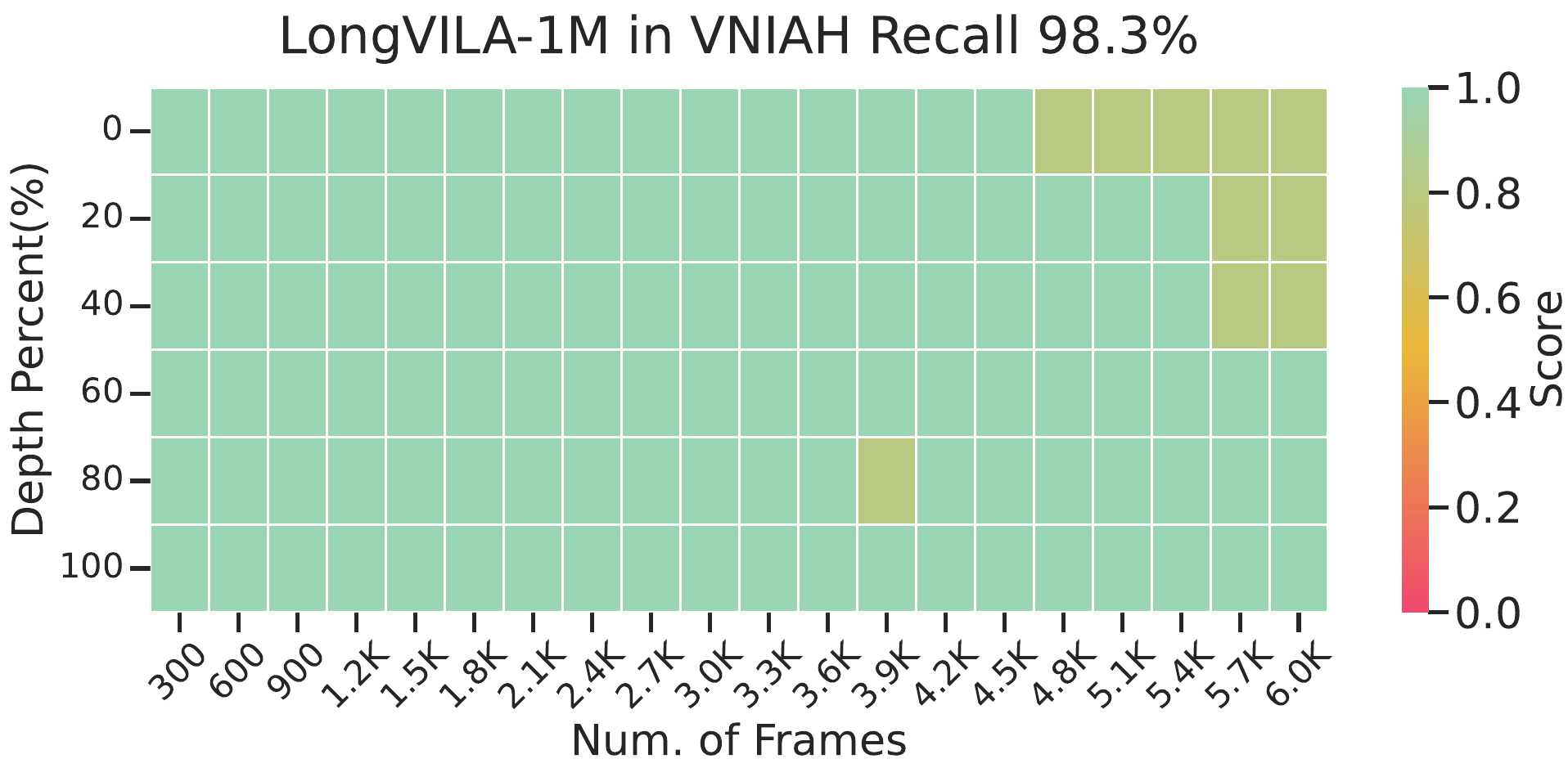}}\\
    \vspace{-2pt}
  \subfloat[\methodall{} in MM-NIAH]{
    \label{sfig:needle_our_mmniah}
    \includegraphics[width=1\columnwidth]{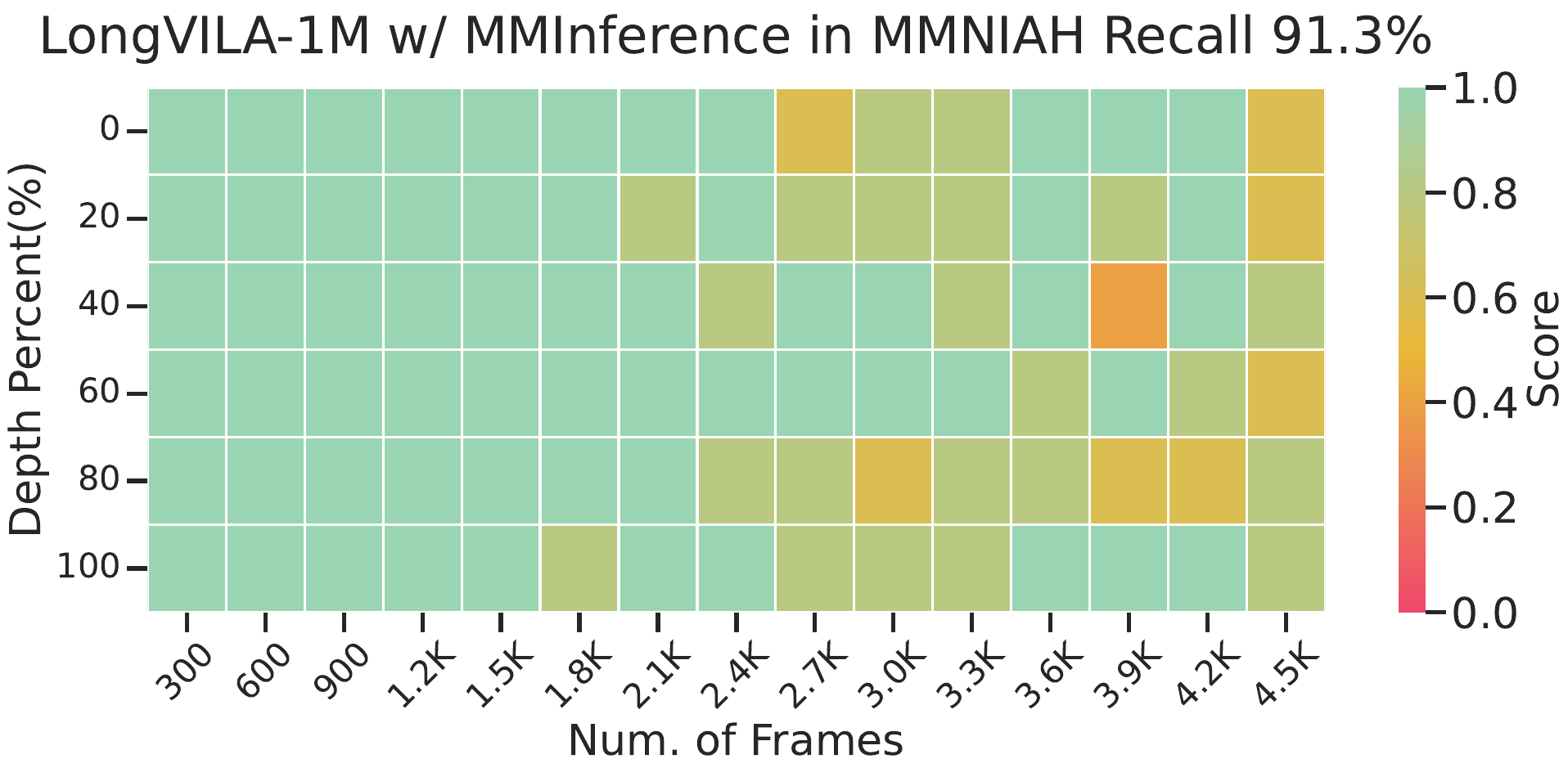}}
  \subfloat[FullAttention in MM-NIAH]{
    \label{sfig:needle_dense_mmniah}
    \includegraphics[width=1\columnwidth]{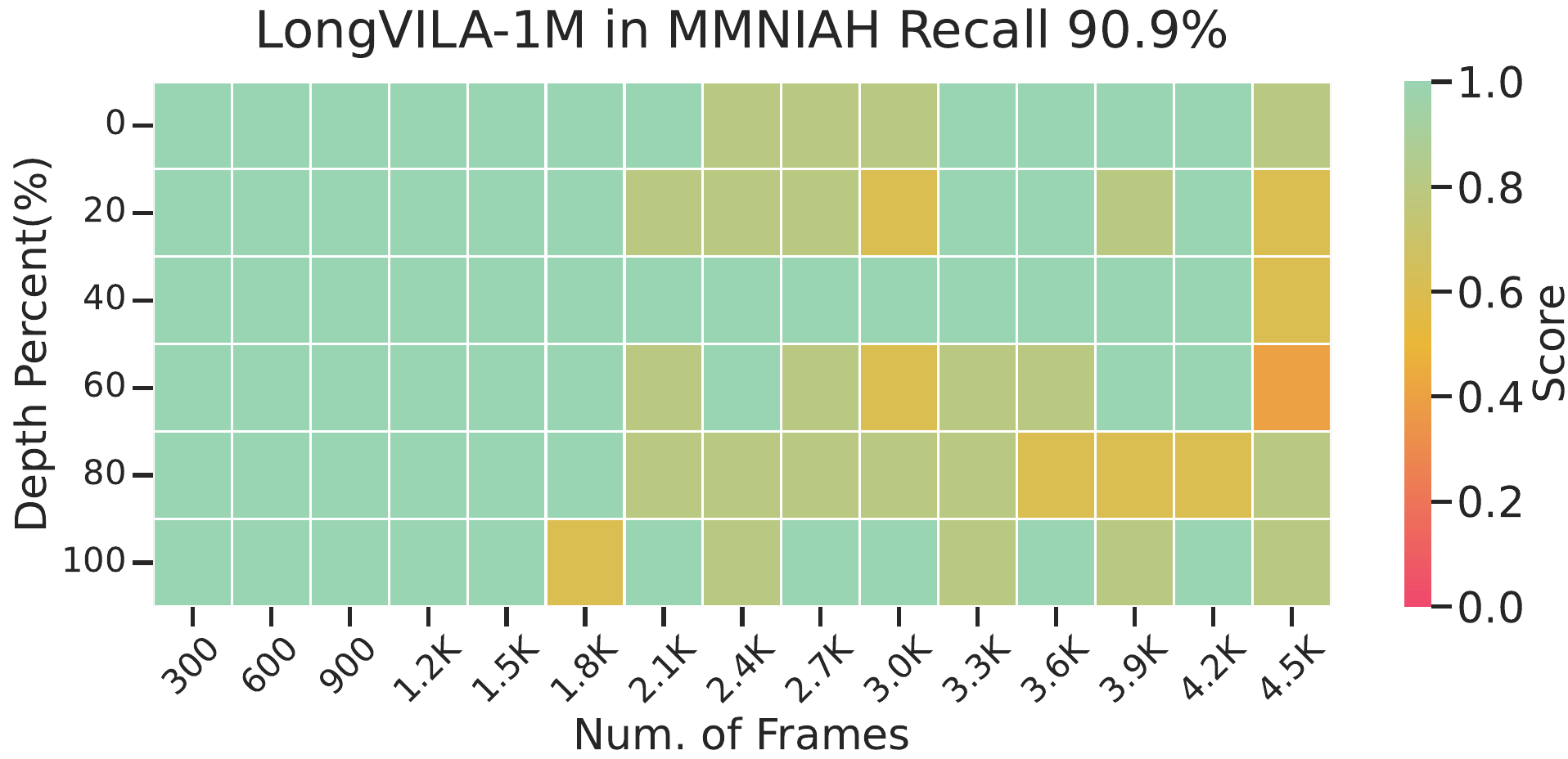}}\\
    \vspace{-5pt}
  \caption{V-NIAH~\cite{zhang2024long} and MM-NIAH results using LongVila-Qwen2-7B-1M~\cite{xue2024longvila}.}
  \label{fig:needle}
\end{figure*}

\paragraph{Implementation Details}

Our experiments are conducted on two state-of-the-art long-video VLMs: Llava-Video~\cite{zhang2024} and LongVILA~\cite{xue2024longvila}. 
We follow the MInference experimental setup, configuring the corresponding search space while adopting optimal configurations from prior work for other methods. We adjust the local window sizes of A-shape and tri-shape patterns to align FLOPs with our method. For MInference, we adopt its optimal configuration, which results with FLOPs approximately twice as high as our method’s in VLMs.
Our implementation leverages Triton~\cite{tillet2019triton}, FlashAttention~\cite{dao2024flashattention2}, and dynamic sparse compiler PIT~\cite{zheng2023pit}.  
For the Vertical-Slash and Grid Head patterns, we set $last_q = 64$. Latency experiments are performed on a single NVIDIA A100 using bfloat16, with greedy decoding to ensure stable results. Additional implementation details are provided in Appendix~\ref{sec:appendix:impl}.

\vspace{-3pt}
\paragraph{Dataset}  
Our evaluation uses the official metrics and scripts provided by these tasks. Additionally, we introduce a Mixed-Modality Needle in a Haystack (MM-NIAH) task to assess VLMs' retrieval capabilities on mixed-modality inputs. Dataset details are provided in Appendix~\ref{sec:appendix:benchmark}.  

{(i) Video Understanding Tasks: These include ActNet-QA~\cite{yu2019}, EgoSchema~\cite{mangalam2023}, Next-QA~\cite{xiao2021}, PerceptionTest~\cite{patraucean2024perception}, VideoDC~\cite{videodc}, and VideoMME~\cite{fu2024}. These benchmarks span five categories, covering tasks such as captioning and video question answering. Input lengths range from 110 frames (e.g., 20k) to 256 frames (e.g., 66k) in Llava-Video~\cite{zhang2024} and LongVILA~\cite{xue2024longvila}.  

{(ii) Video Needle in a Haystack (V-NIAH)~\cite{zhang2024long}:} A long-video retrieval task testing VLMs' performance with tokens of up to 6k frames (e.g., 1.1M tokens), where inserted images are placed at various positions.  

{(iii) Mixed-Modality Needle in a Haystack (MM-NIAH):} To evaluate VLMs in mixed-modality scenarios, we construct a mix-modality version of NIAH. Specifically, 25\% of the input consists of text segments inserted at the document level across different frames in long-video inputs, forming a mix-modality haystack. All other settings align with V-NIAH, including the multi-choice VQA task with randomly inserted images. This benchmark tests input lengths of up to 4.5k frames (e.g., 1.1M tokens).

\begin{figure*}[htb]
    \vspace{-10pt}
    \centering
    \subfloat[All Textual Context]{
        \includegraphics[height=0.45\columnwidth]{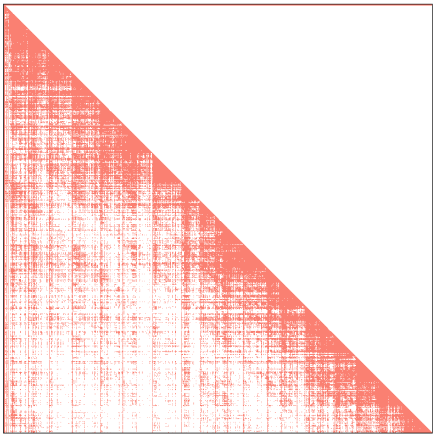}
    }
    \hspace{0.2em}
    \subfloat[Visual Context Inserted]{
        \includegraphics[height=0.45\columnwidth]{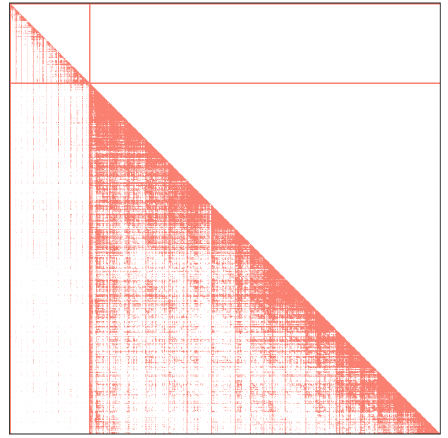}
    }
    \hspace{0.2em}
    \subfloat[More Visual Context]{
        \includegraphics[height=0.45\columnwidth]{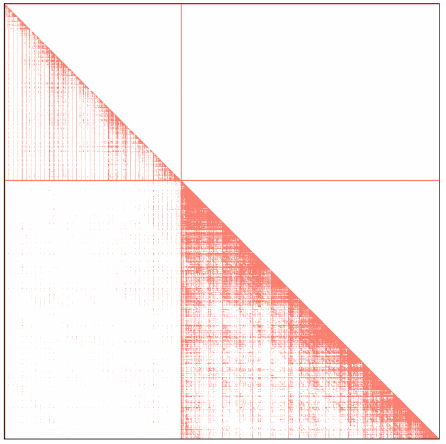}
    }
    \hspace{0.2em}
    \subfloat[All Visual Context]{
        \includegraphics[height=0.45\columnwidth]{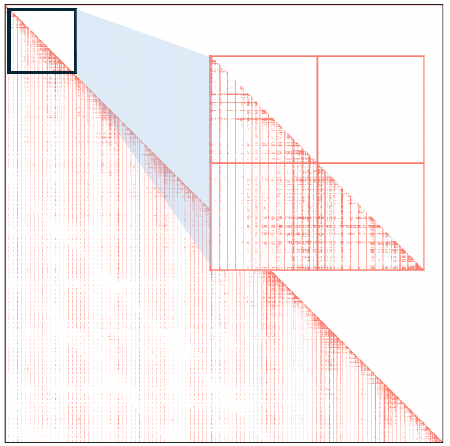}
    }
    \caption{Transition of sparse patterns from textual context to visual context. (a) The vertical-slash pattern for all textual context. (b) Grid pattern appears when visual modality is appended. (c) Grid pattern dominates.}
    \label{fig:modality-transition}
\vspace{-10pt}
\end{figure*}

\paragraph{Baselines} 
We include five training-free sparse attention approaches, one visual token compression method, and also incorporate FlashAttention-2~\cite{dao2024flashattention2} as a baseline.
1) SparseTransformer (Fixed)~\cite{child2019}: Retains attention within each segment and allows all tokens to attend to the segment’s initial tokens.
2) SparseTransformer (Strided)~\cite{child2019}: Employs local windows with dilated attention.  
3) A-Shape~\cite{xiao2024efficient}: Preserves only the sink token with local attention.
4) Tri-Shape~\cite{li2024scbench,acharya2024star}: Extends A-Shape by enabling full attention for all tokens to the last window’s queries.
5) Vertical-Slash Pattern~\cite{jiang2024minference}: Focuses on specific tokens (vertical lines) and tokens at fixed intervals (slash lines).
6) VisionZip~\cite{yang2024visionzip}: A visual token compression method that reduces the number of visual tokens per frame by evaluating tokens based on their attention scores and discarding less important ones.
Full details on implementation, hyperparameters, and illustrations for our baselines can be found in Appendix \ref{sec:appendix:impl}.

\subsection{Long Video Understanding}

Table~\ref{tab:benchmark_results} presents the performance of different methods on video understanding tasks. The results show that:
1) Our method and MInference closely approximate full attention across all tasks while requiring only half the FLOPs of MInference.
2) Static sparse patterns, such as A-shape and Tri-shape, perform reasonably well on most tasks but experience a notable performance drop in multi-choice VQA tasks like EgoSchema. Additionally, the slight increase in query full attention in Tri-shape effectively improves performance.
3) Among SF patterns, the slash pattern better preserves performance. Even when using SF-fixed with only 2\%-5\% of FLOPs, it still maintains strong performance on most tasks.

\subsection{Video Needle In A Haystack}

Fig.~\ref{sfig:needle_our_vniah}, \ref{sfig:needle_dense_vniah}, and \ref{fig:needle_vniah} show the performance of different models on V-NIAH, revealing notable differences in handling long-context video retrieval as the number of processed frames increases:
1) Our method achieves results nearly identical to full attention.
2) A-shape struggles with mid-context information even at 300 frames, while Tri-shape maintains full performance until 3.9k frames (i.g. 700K tokens) before a sharp decline.
3) SF-fixed degrades at 2.1k frames (i.g. 350K tokens), while SF-strided surpasses Tri-shape, holding performance until 4.5k frames (i.g. 825K tokens).
4) MInference preserves VLM retrieval well, with only slight degradation beyond 4.8K frames.

\subsection{Mixed-Modality Needle In A Haystack}

Beyond V-NIAH, we introduce a mixed-modality NIAH test to evaluate the performance of different sparse methods on video-text inputs, in Fig.~\ref{sfig:needle_our_mmniah}, \ref{sfig:needle_dense_mmniah}, and \ref{fig:needle_mmniah}.
Mixed-modality inputs lead to more pronounced performance degradation across all methods. However, by incorporating inter-modality sparse patterns, our method maintains performance close to full attention, especially when compared to MInference and ours w/o inter-modality. Notably, Tri-shape and MInference show significant drops at 1.8k frames (i.g. 440K tokens) and 2.7k frames (i.g. 660K tokens).

\subsection{Latency}

\begin{figure}[htb]
    \vspace{-10pt}
    \centering
    \includegraphics[height=0.53\columnwidth]{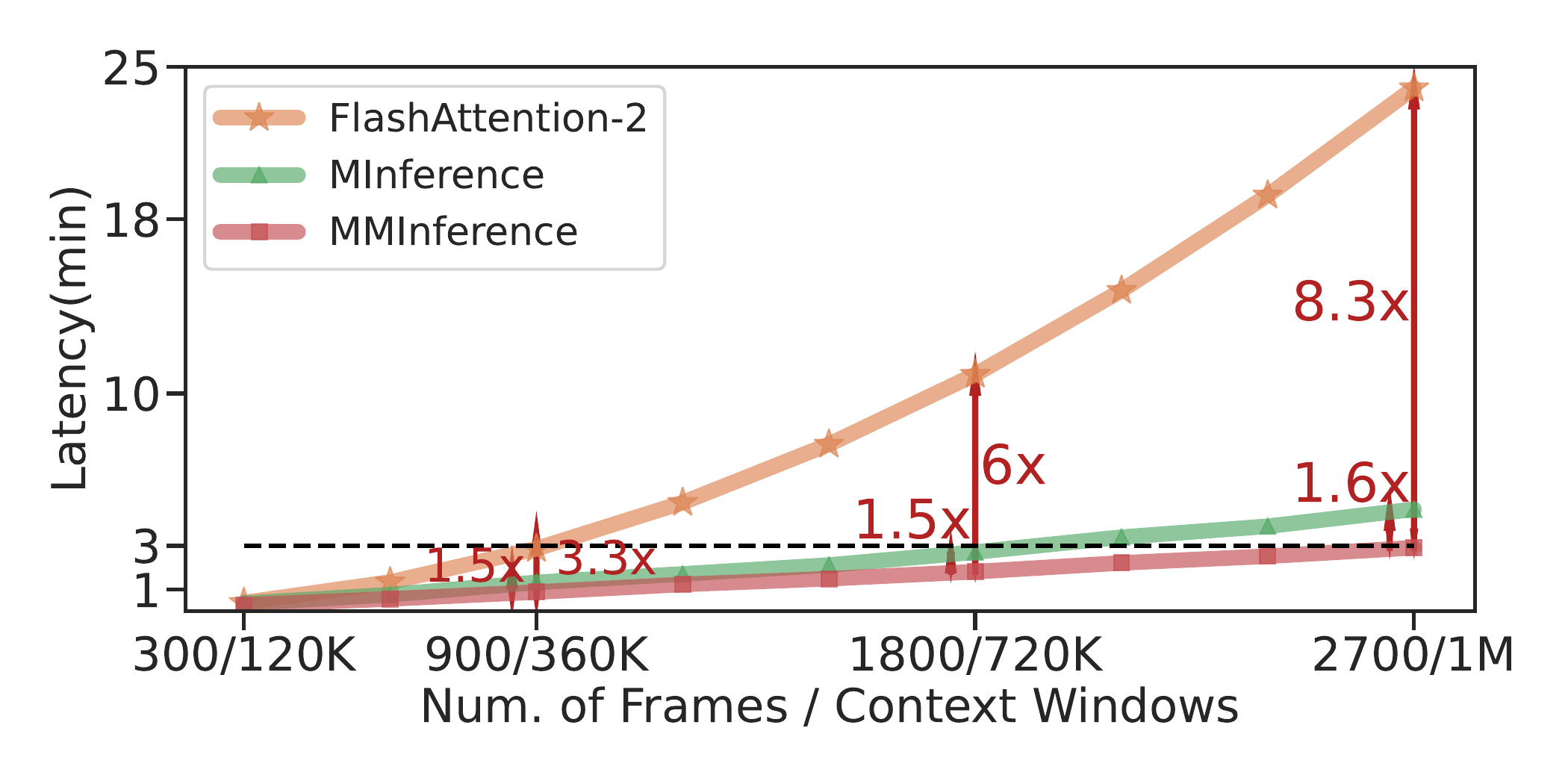}
    \caption{End-to-End Latency.}
    \label{fig:latency}
    \vspace{-10pt}
  \end{figure}
  
Fig.~\ref{fig:latency} and \ref{fig:latency_breakdown_detail} present end-to-end and kernel-level latency across different context sizes. The grid pattern significantly outperforms the vertical-slash pattern in sparsity, achieving a 2–3× speedup even at 1M tokens. Additionally, the grid pattern achieves an end-to-end speedup of 8.3× and a kernel-level speedup of 12×.

\subsection{Analysis}
\label{subsec:analysis}

\begin{table*}[t]
    \centering
    \small
    \caption{Performance (\%) on video understanding tasks based on VideoChat-Flash~\cite{li2024videochat} at frames 512 with 8k tokens.}
    \label{tab:videochat}
    \resizebox{0.9\textwidth}{!}{
    \begin{tabular}{l c c c c c c c c}
        \toprule
        \multirow{2}{*}{\textbf{Model}} & \textbf{VideoDC} & \textbf{ActNet-QA} & \textbf{EgoSchema} & \textbf{Next-QA} & \textbf{PerceptionTest} & \multicolumn{2}{c}{\textbf{VideoMME}} & \multirow{2}{*}{\textbf{Avg.}} \\
        \cmidrule(lr){2-8}
        & test & test & test & mc & val & w/o sub. & w/ sub. & \\
        \midrule
        \rowcolor{Gray}
        VideoChat-Flash & 3.21 & 53.6 & 57.0 & 81.2 & 69.1 & 63.2 & 70.5 & 56.8 \\
        w/ \methodall{} & 3.19 & 54.3 & 57.3 & 79.8 & 69.1 & 63.0 & 70.2 & 56.7 \\
        \bottomrule
    \end{tabular}
    }
\end{table*}

\paragraph{Transition of Sparse Patterns Across Modalities}

Since LLMs and VLMs exhibit different sparse patterns, we examine the interplay between the Grid and  Vertical-Slash pattern. As shown in Fig.~\ref{fig:modality-transition}, Llava-Video-7B primarily uses Vertical-Slash pattern for purely textual inputs. However, once a visual input is appended, it transitions to a Grid pattern to capture the geometric structure of the visual content. This shift occurs at the modality boundary, creating a more structured arrangement of vertical and horizontal intervals. Such behavior highlights the need for distinct sparsity strategies in visual and mixed-modality contexts, rather than simply reusing sparse patterns from LLMs for VLMs.

\begin{figure}[htb]
    \vspace{-5pt}
    \centering
    \includegraphics[height=0.58\columnwidth]{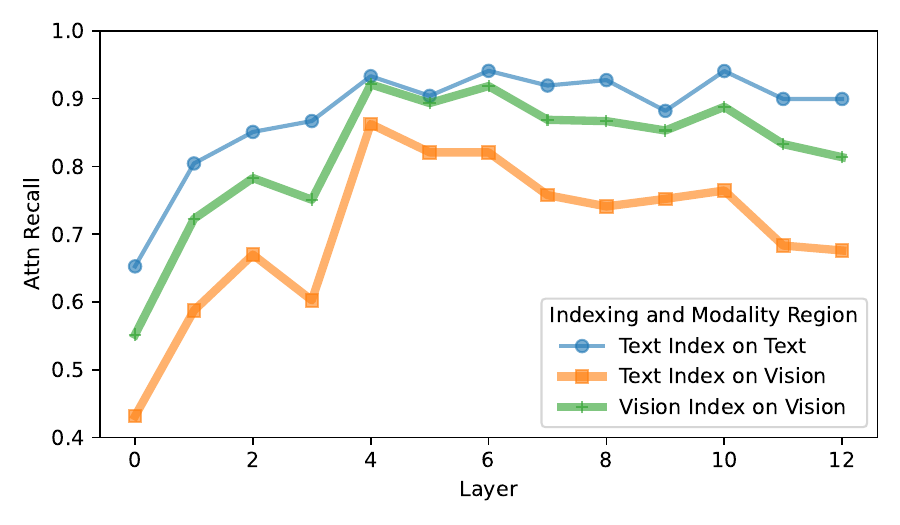}
    \vspace{-10pt}
    \caption{The sparse index does not effectively extrapolate from text to the visual modality. However, an index built within the same modality can generalize across modality boundaries.}
    \label{fig:continuity}
    \vspace{-10pt}
  \end{figure}
  
\paragraph{Sparse Index Across Modalities}
In Fig.~\ref{fig:continuity}, the sparse index achieves high recall for textual regions but fails to generalize to visual ones. To address this, we construct a sparse index from the visual modality and evaluate it on separate visual segments, each separated by modality boundaries. Remarkably, this approach extrapolates effectively across all visual segments, even when interspersed with textual boundaries. As shown in Fig.~\ref{fig:continuity}, the sparse index achieves high recall in the textual but fails to generalize to the visual. To address this, we construct a sparse index using the visual modality and evaluate it across distinct regions of the visual modality, separated by modality boundaries. Remarkably, this approach successfully extrapolates to all visual regions even when interrupted by text-induced boundaries.

\paragraph{Integrate with token compression methods}

As shown in Table~\ref{tab:videochat}, our method integrates seamlessly with token compression techniques, enabling near-lossless performance while supporting longer or higher-resolution video inputs. Specifically, VideoChat-Flash reduces tokens per frame from 196 to 16 at the ViT stage, while our method further applies sparse attention in the LLM decoder. Results demonstrate strong performance retention across benchmarks.

\section{Related Work}

\paragraph{Long-Context Vision Language Models} Recent VLMs have extended their context length to support long multi-modal inputs~\cite{zhang2024long,xue2024longvila,wang2024longllava,reid2024gemini}, enabling applications such as long-video understanding~\cite{fu2024,xiao2021,wang2024qwen2vl,bai2025qwen2}, multi-modal retrieval~\cite{zhang2024long}, and multi-modal chain-of-thought reasoning~\cite{qvq-72b-preview}. For instance, \citet{zhang2024long} transfer long-context capabilities from base LLMs to vision tasks, \citet{xue2024longvila} introduce multi-modal sequence parallelism to accelerate video fine-tuning, and \citet{zhang2024} emphasize the role of data calibration and synthetic data in boosting VLM performance.

\paragraph{Efficiency Optimization for VLMs} While long-context VLMs achieve high accuracy, their high inference cost limits practical use in long-video scenarios. A common strategy is \textit{vision token compression}—reducing video feature resolution by dropping or merging less important visual tokens \cite{bolya2023token,chen2024an,shen2024longvu,he2024zipvl,tu2024vlcache,weng2024longvlm,wen2024efficient}. RNN-Transformer hybrids are also used \cite{wang2024longllava} to balance efficiency and context length. However, these methods often assume inputs are long videos paired with short text, focusing solely on visual token optimization, while overlooking mixed-modality inputs critical for multi-turn interactions \cite{huang2024dialoggen}.
Recently, \citet{xu2025xattention} applied dynamic sparse attention to long-context VLMs, but their approach ignores modality-specific inductive biases and is limited to single-modality video tasks.

\section{Conclusion}

We propose \methodall{}, a modality-aware permutation sparse attention method that accelerates long-context VLMs. It features permutation-based grid sparse attention, Q-boundary/2D-boundary patterns for mixed-modality boundaries, and a Modality-Aware Sparse Attention Search Algorithm. Our optimized GPU kernels enable end-to-end acceleration. Experiments on video understanding tasks, V-NIAH and MM-NIAH using Llava-Video and LongVila demonstrate that \methodall{} preserves full-attention performance while achieving up to 8.3× speedup at 1M tokens.

\section*{Impact Statement}

This paper presents work whose goal is to advance the field of Machine Learning. There are many potential societal consequences of our work, none which we feel must be specifically highlighted here.

\bibliography{ref}
\bibliographystyle{icml2025}

\newpage
\appendix
\onecolumn

\section{Modality-Aware Sparse Attention Search Algorithm}

In Algorithm~\ref{alg:search}, we detail the procedure for selecting the optimal sparse attention pattern for each attention head under a constrained FLOPs budget. The algorithm jointly determines the best pattern and its configuration (e.g., stride size in grid attention, number of vertical/slash lines in VS pattern) to maximize accuracy. We first construct a kernel-aware search space, where all candidate patterns have comparable real-world FLOPs based on GPU kernel measurements—rather than theoretical estimates—to ensure practical efficiency.

We then evaluate each candidate using a reference example and select the configuration that maximizes attention recall, using the actual attention output as the objective. This recall-based scoring incorporates the V matrix and builds on FlashAttention~\cite{dao2024flashattention2}, enabling end-to-end pattern selection with minimal memory overhead and improved performance.

\restylefloat{algorithm}

\begin{figure}[htb]
  \centering
\begin{minipage}[t]{0.6\textwidth}
\begin{algorithm}[H]
\captionsetup[algorithm]{singlelinecheck=off}
\caption{Modality-aware Sparse Attention Pattern Search}
\label{alg:search}
\begin{algorithmic}
  \STATE {\bfseries Input:} $\boldsymbol{Q},\boldsymbol{K},\boldsymbol{V} \in \mathbb{R}^{S \times d_h}$, inter-modality search space $\rho_{\mathrm{inter}}$, intra-modality search space $\rho_{\mathrm{intra}}$, modality type set $m\in \phi_m$, optimized sparse pattern ${\mathrm{P}}$

    \LineComment{Intra-modality sparse attention pattern search}
    \FOR{$i \gets 1$ to $|\phi_m|$}
    \STATE $p_{mi} \gets \mathrm{KernelAwareSearch}\left(\boldsymbol{Q}, \boldsymbol{K}, \boldsymbol{V}, m_i\right)$
    \STATE $\mathrm{P} \gets \mathrm{P} \cup p_{mi}$
    \ENDFOR

    \LineComment{Cross-modality sparse attention pattern search}
    \FOR{$i \gets 1$ to $|\phi_m|$}
    \FOR{$j \gets 1$ to $|\phi_m|$}
    \STATE $p_{mi,mj} \gets \mathrm{KernelAwareSearch}\left(\boldsymbol{Q}, \boldsymbol{K}, \boldsymbol{V}, m_i, mj\right)$
    \STATE $\mathrm{P} \gets \mathrm{P} \cup p_{mi, mj}$
    \ENDFOR
    \ENDFOR
 
    \LineComment{Inter-modality sparse attention pattern search}
    \FOR{$i \gets 1$ to $|\rho_{\mathrm{inter}}|$}
        \STATE $p_{i} \gets \mathrm{argmin}\left(|\mathrm{sparse}(\boldsymbol{{Q}}, \boldsymbol{{K}}, \boldsymbol{{V}}, i\right)-$
        \STATE $\mathrm{attention}(\boldsymbol{{Q}}, \boldsymbol{{K}}, \boldsymbol{{V}})|$
        \STATE $\mathrm{P} \gets \mathrm{P} \cup p_{i}$
    \ENDFOR

    \STATE $\mathrm{return}\,\,\,\mathrm{P}$
   
\end{algorithmic}
\end{algorithm}
\end{minipage}
\end{figure}

\begin{figure*}[htb]
    \centering
    \subfloat[K-Boundary pattern.]{
        \label{sfig:k_boundary_pattern}
        \includegraphics[height=0.3\columnwidth]{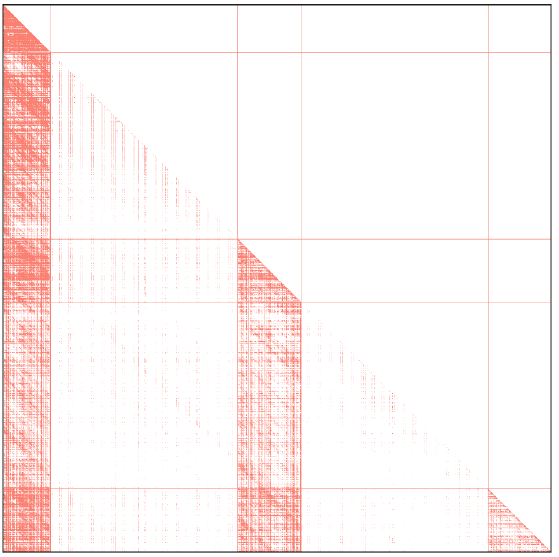}}
    \subfloat[No-Boundary pattern.]{
        \label{sfig:no_boundary_pattern}
        \includegraphics[height=0.3\columnwidth]{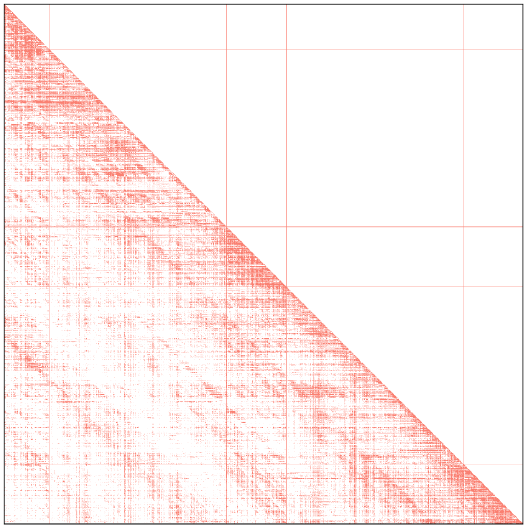}}
    \caption{Additional inter-modality sparse pattern.}
    \label{fig:add_boundary_pattern}
\end{figure*}

\section{Pattern Analysis}

\subsection{Additional Mix-modality Pattern}

In \S\ref{sec:motivation}, we explain how the grid pattern naturally arises from the geometric structure of vision inputs. Fig.~\ref{fig:add_boundary_pattern} further illustrates two additional patterns in the mixed-modality search space: the K-Boundary and No-Boundary patterns. Notably, both patterns incur no additional cost compared to pure intra-modality attention, as their sparse indices can be computed across all rows without extra computation.

\subsection{Additional Sparse Attention Pattern Visualization}

We further analyze the sparse patterns in Qwen2.5-VL~\cite{wang2024qwen2vl} with dynamic resolution inputs and in VideoChat-Flash~\cite{li2024videochat} under visual token compression, across both video benchmark and mixed-modality inputs, as shown in Fig.\ref{fig:additional_pattern_viz} and Fig.\ref{fig:additional_pattern_viz_mix}.

\section{Experiment Details}
\label{sec:appendix:impl}

\subsection{Vision Language Models}

We use two state-of-the-art VLMs in our experiments: LongVILA~\cite{xue2024longvila} and Llava-Video~\cite{zhang2024}. Llava-Video supports varying numbers of frames (32, 64, 110) for video understanding, and as reported, performance improves with more frames. Thus, we adopt the 110-frame variant for benchmarking. For LongVILA, we use the 256-frame version (LongVILA-256Frame) with a 128K context length for video understanding benchmarks, and the 1M-token version (LongVILA-1M), designed for retrieval tasks, for the V-NIAH evaluation.

\subsection{Baselines}

\begin{figure*}
    \centering
    \subfloat[A-shape]{
        \label{sfig:a_shape}
        \includegraphics[height=0.25\columnwidth]{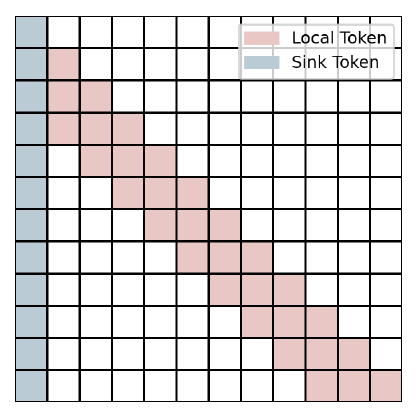}}
    \subfloat[SF-fixed]{
        \label{sfig:sf_fixed}
        \includegraphics[height=0.25\columnwidth]{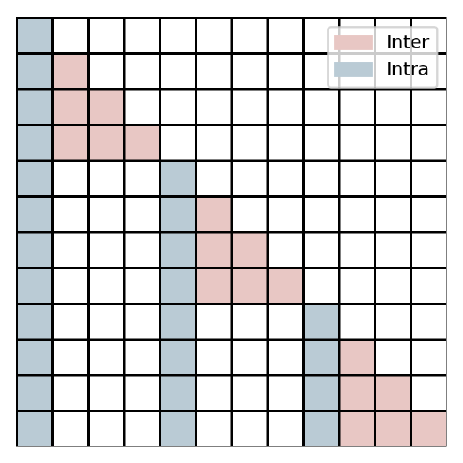}}
    \subfloat[SF-strided]{
        \label{sfig:sf_strided}
        \includegraphics[height=0.25\columnwidth]{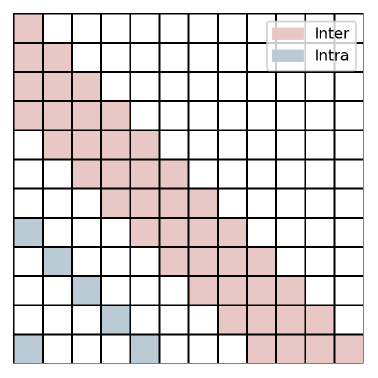}}
    \\
    \subfloat[Tri-shape]{
        \label{sfig:tri_shape}
        \includegraphics[height=0.25\columnwidth]{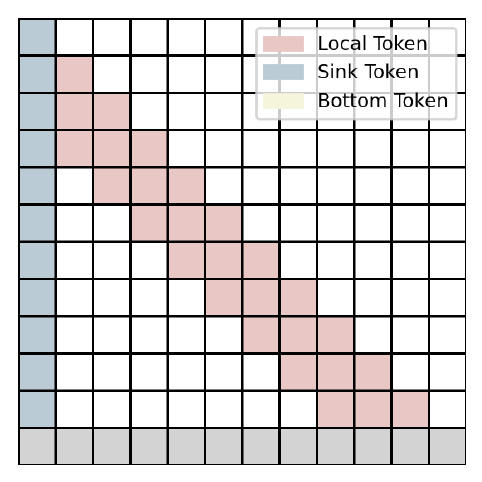}}
    \subfloat[Vertical-Slash (MInference)]{
        \label{sfig:minference}
        \includegraphics[height=0.25\columnwidth]{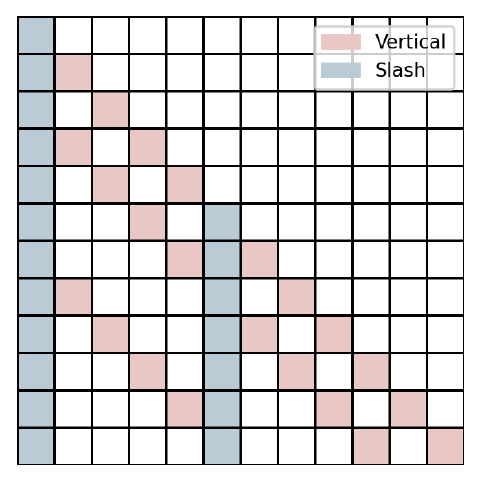}}
    \caption{The baselines of sparse attention in our experiments.}
    \label{fig:baselines}
\end{figure*}

We include five sparse attention baselines in our experiments: A-shape~\cite{xiao2024efficient}, SF-fixed~\cite{child2019}, SF-strided~\cite{child2019}, Tri-shape~\cite{li2024scbench}, MInference~\cite{jiang2024minference}, and VisionZip~\cite{yang2024visionzip}. Fig.~\ref{fig:baselines} illustrates the attention patterns of these baselines.

While VisionZip~\cite{yang2024visionzip} is primarily a visual token compression method—compressing vision tokens using attention scores from the vision encoder before passing them to the LLM—it is included for comparison as it reduces FLOPs in the pre-filling stage and offers insight into token compression approaches.

\begin{table}[t]
    \centering
    \caption{Hyperparameters detail of baselines.}
    \label{tab:impl_details}
    \resizebox{0.75\columnwidth}{!}{
    \begin{tabular}{l|l}
    \toprule
    Method & Hyperparameters \\
    \midrule
    A-shape & $\mathrm{Sink}=128, \mathrm{Local}=4096$ \\
    SF-fixed & $\mathrm{Local}=\mathrm{token\_per\_frame}, \mathrm{vline\_stride}=\mathrm{token\_per\_frame}$ \\
    SF-strided & $\mathrm{Local}=\mathrm{token\_per\_frame}, \mathrm{vline\_stride}=\mathrm{token\_per\_frame}$ \\
    Tri-shape & $\mathrm{Sink}=128, \mathrm{Local}=4096, \mathrm{Bottom}=128$ \\
    MInference & $\mathrm{Vertical\_size} \in \{1000, 2000, 4000\}, \mathrm{Slash\_size} \in \{1024, 2048, 4096, 6144\}$ \\
    VisionZip & $\mathrm{dominant} = 54, \mathrm{contextual} = 10$ \\
    \bottomrule
    \end{tabular}
    }
\end{table}

\subsection{A-shape and Vertical-Slash }
\label{sec:appendix:a_shape_vs}

A-shape and Vertical-Slash are used for intra-modality attention, alongside our newly proposed Grid pattern.

At inference time, we estimate the attention matrix online to dynamically determine the spatial layout of sparse indices, conditioned on the assigned pattern and actual input. Sparse attention is then computed using our optimized GPU kernels. Note that while the masks for Vertical-Slash and Grid patterns are dynamically generated, A-shape uses a static mask, incurring no additional overhead beyond sparse computation.

\textit{A-shape head.}
A-shape is a static sparse pattern that includes the first seven initial tokens along with a local attention window.

\textit{\textit{Vertical-Slash} head.}
Due to the continuity of vertical and slash lines, we matmul the last query vector $\bm{Q}_{[-\text{last\_q}:]}$ and key vector $\bm{K}$ to produce the estimated attention matrix $\boldsymbol{\hat{A}}$, which, in turn, is used to determine the indices for the vertical $\bm{i}_v$ and slash $\bm{i}_s$ lines.
After obtaining the sparse indices for the vertical and slash lines, we convert them into a sparse format $\bm{i}_{vs}$. Using these sparse indices, we perform block-sparse calculations of the attention weights and attention output.

\subsection{Permutation for the Grid Pattern and Across Modality}

We illustrate how the permutation is applied to the Grid pattern and the Q-boundary and 2D-boundary patterns in Fig.~\ref{fig:grid_pattern_permutation} and Fig.~\ref{fig:grid_pattern_permutation}.

\begin{figure*}[htb]
    \centering
    \subfloat[Before Permutation]{
        \label{sfig:grid_pattern_before_permutation}
        \includegraphics[height=0.25\columnwidth]{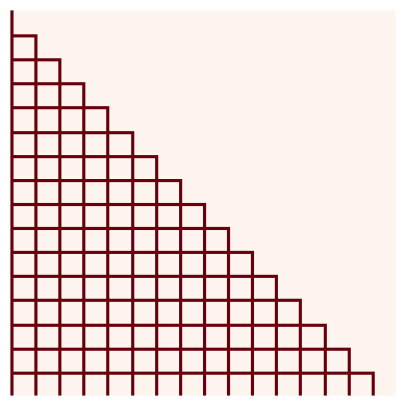}}
    \subfloat[Row-wise Permutation]{
        \label{sfig:grid_pattern_permutation_q}
        \includegraphics[height=0.25\columnwidth]{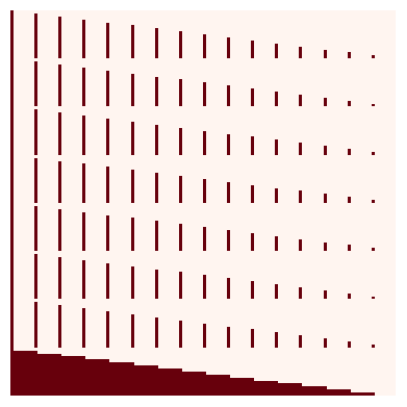}}
    \subfloat[Column-wise Permutation]{
        \label{sfig:grid_pattern_permutation_k}
        \includegraphics[height=0.25\columnwidth]{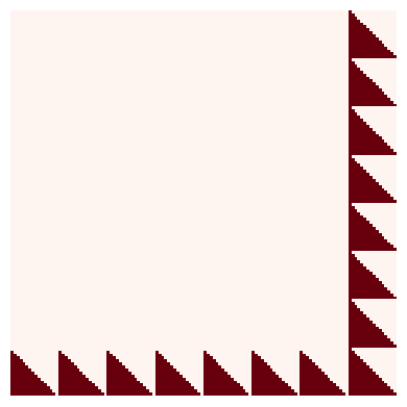}}
    \caption{Permutation for the Grid Pattern. (a) Before permutation. (b) Row-wise permutation. (c) Column-wise permutation.}
    \label{fig:grid_pattern_permutation}
\end{figure*}

\begin{figure*}[htb]
    \centering
    \subfloat[Mix-modality]{
        \label{sfig:mix_modality}
        \includegraphics[height=0.25\columnwidth]{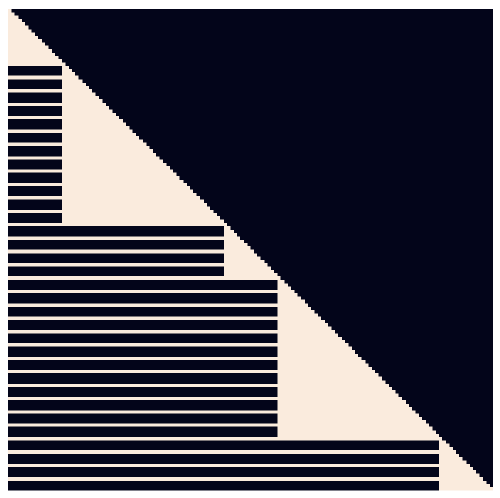}}
    \subfloat[Q-wise Permutation]{
        \label{sfig:q_wise_permutation}
        \includegraphics[height=0.25\columnwidth]{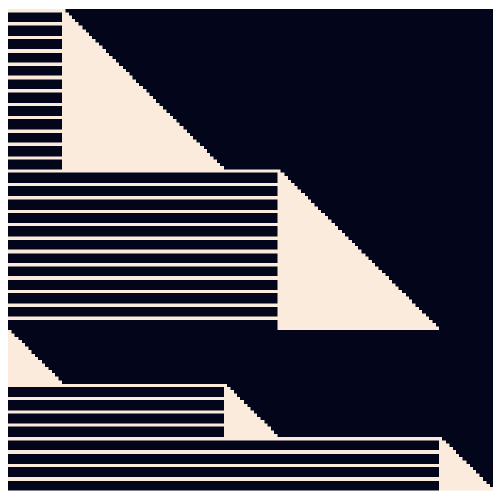}}
    \subfloat[K-wise Permutation]{
        \label{sfig:k_wise_permutation}
        \includegraphics[height=0.25\columnwidth]{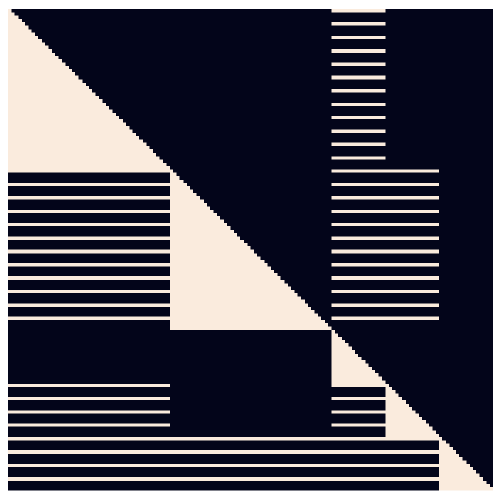}}
    \caption{Permutation for mix-modality context. (a) Mix-modality. (b) Q-wise permutation. (c) K-wise permutation.}
    \label{fig:mix_modality_permutation}
\end{figure*}

\subsection{Search Space}

Following \cite{jiang2024minference}, we set the target FLOPs $t$ to be the same as 1k global tokens and 4k local window tokens in the \textit{A-shape} pattern. Additionally, we use only one sample as our calibration set from the egoschema task with no more than 25K tokens, which exhibits strong generalization and stability across different lengths and domains. The search time is approximately 15 minutes on a single A100. This pattern search is individually conducted for each model: Llava-Video-7B, LongVila-256Frame, and LongVila-1M. The search space is shown in Table~\ref{tab:search_space}.

\begin{table*}[t]
    \centering
    \begin{tabular}{ll}
    \toprule
    \textbf{Attention Type} & \textbf{Parameters} \\
    \midrule
    \multirow{12}{*}{Grid Attention} & (frame\_stride, True, False, False, 1024) \\
    & (frame\_stride, False, True, False, 1024) \\
    & (frame\_stride, False, False, True, 1024) \\
    & (frame\_stride, True, True, False, 1024) \\
    & (frame\_stride, False, True, True, 1024) \\
    & (frame\_stride, True, True, True, 1024) \\
    & (stride, True, False, False, 1024) \\
    & (stride, False, True, False, 1024) \\
    & (stride, False, False, True, 1024) \\
    & (stride, True, True, False, 1024) \\
    & (stride, False, True, True, 1024) \\
    & (stride, True, True, True, 1024) \\
    \midrule
    \multirow{3}{*}{A-shape} & (128, 1024) \\
    & (128, 2048) \\
    & (128, 4096) \\
    \midrule
    \multirow{10}{*}{Vertical-Slash} & (1000, 1024) \\
    & (1000, 2048) \\
    & (2000, 2048) \\
    & (1000, 3096) \\
    & (2000, 3096) \\
    & (1000, 4096) \\
    & (2000, 4096) \\
    & (3500, 200) \\
    & (1000, 2500) \\
    \bottomrule
    \end{tabular}
    \caption{The search space for each attention pattern: 1) Grid Attention: (stride, use hline, use vline, use slash, max stride); 2) A-shape: (sink, local); 3) Vertical-Slash: (vertical size, slash size)}
    \label{tab:search_space}
\end{table*}

\section{Benchmark Details}
\label{sec:appendix:benchmark}

We evaluate our method on several video understanding benchmarks that test different aspects of video comprehension:

\paragraph{EgoSchema}
EgoSchema \cite{mangalam2023} is a diagnostic benchmark for very long-form video language understanding, structured as a multiple-choice question answering task. The benchmark requires models to answer questions about egocentric videos by selecting from given options (labeled A through E). The evaluation can be performed either on the full set via submission to an evaluation server, or on a released subset of 500 questions for direct scoring.

\paragraph{Video-MME}
Video-MME \cite{fu2024} is a comprehensive multi-modal evaluation benchmark that tests MLLMs across diverse video types and temporal dimensions. It spans 6 primary visual domains with 30 subfields and includes videos ranging from 11 seconds to 1 hour in duration. The benchmark comprises 900 videos totaling 254 hours, with 2,700 manually annotated question-answer pairs. It evaluates models' ability to process not just video frames but also integrated multi-modal inputs like subtitles and audio.

\paragraph{NExT-QA}
NExT-QA \cite{xiao2021} focuses on advancing video understanding from basic description to explaining temporal actions. It features both multiple-choice and open-ended QA tasks that target three key aspects: causal action reasoning, temporal action reasoning, and common scene comprehension. The benchmark is specifically designed to evaluate models' ability to reason about actions beyond superficial scene descriptions.

\paragraph{Perception Test}
The Perception Test \cite{patraucean2023perception} perce evaluates perception and reasoning skills across video, audio, and text modalities. It contains 11.6k real-world videos with an average length of 23 seconds, featuring perceptually interesting situations. The benchmark tests four key skills (Memory, Abstraction, Physics, Semantics) and various types of reasoning (descriptive, explanatory, predictive, counterfactual). Videos are densely annotated with six types of labels: multiple-choice QA, grounded video QA, object tracks, point tracks, temporal action segments, and sound segments.

\paragraph{ActivityNet-QA}
ActivityNet-QA \cite{yu2019} is a large-scale VideoQA dataset consisting of 58,000 QA pairs on 5,800 complex web videos derived from the ActivityNet dataset. The benchmark is fully annotated and designed to test models' understanding of complex web videos through question answering. Unlike automatically generated datasets, ActivityNet-QA features human-annotated questions and answers, making it particularly valuable for evaluating real-world video understanding capabilities.

\paragraph{Video Detail Description (VideoDC)}
VideoDC \cite{videodc} focuses on comprehensive video understanding through detailed descriptions. The benchmark consists of question-answer pairs generated with GPT-3.5, where questions prompt for detailed descriptions focusing on main subjects, their actions, and background scenes. The evaluation assesses the quality and completeness of video descriptions generated by models.

\section{Additional Experiments Results}

\subsection{Additional Video Needle In A Haystack Results}

we further present the results of the Video Needle In A Haystack task with our baselines. The results of our method and full atttenton is shown in Fig.~\ref{fig:needle}.

\begin{figure*}[htb]
  \centering
  \subfloat[A-shape]{
    \label{sfig:needle_a_shape_vniah}
    \includegraphics[width=0.5\columnwidth]{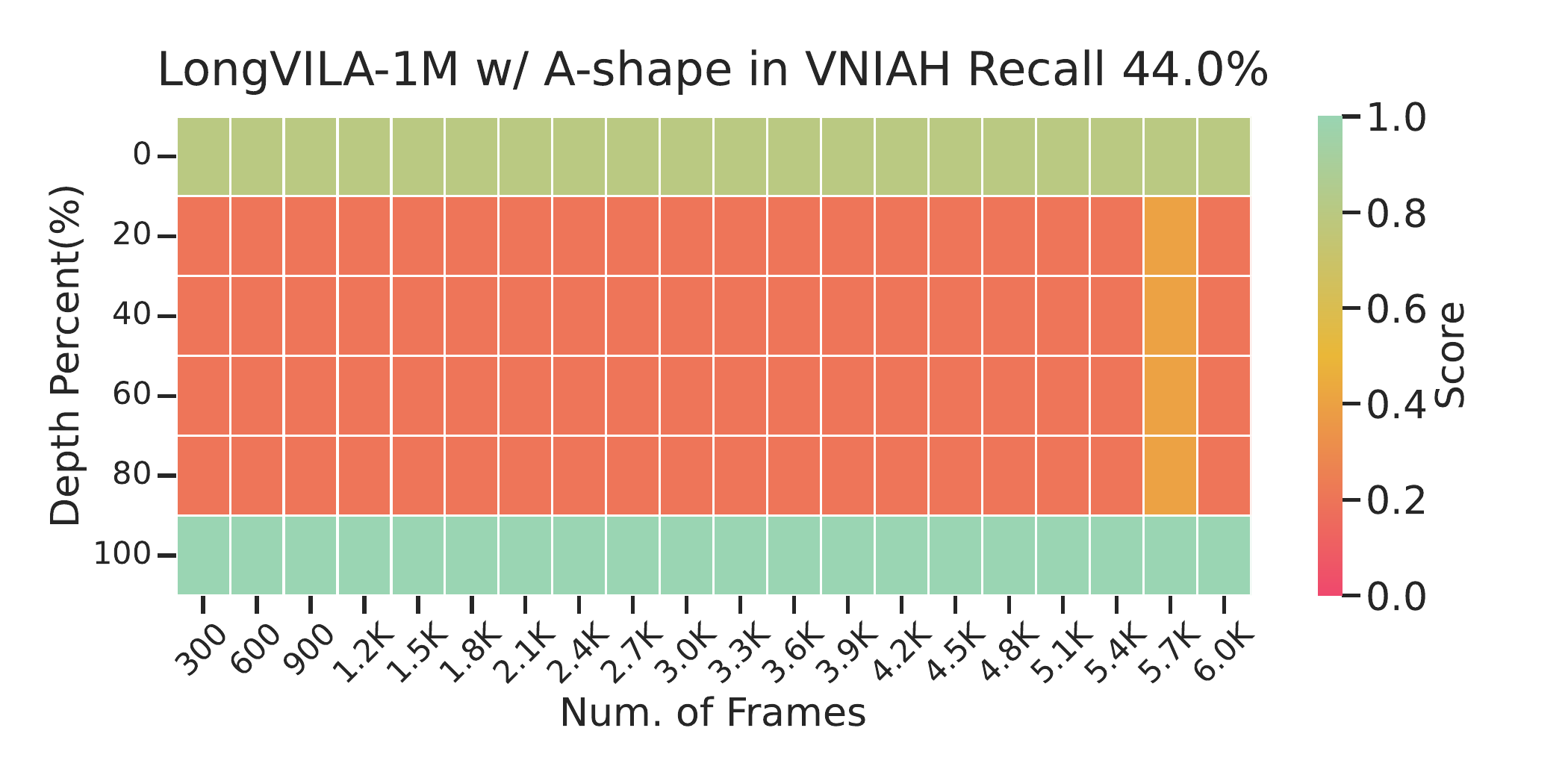}}
  \subfloat[Tri-shape]{
    \label{sfig:needle_tri_shape_vniah}
    \includegraphics[width=0.5\columnwidth]{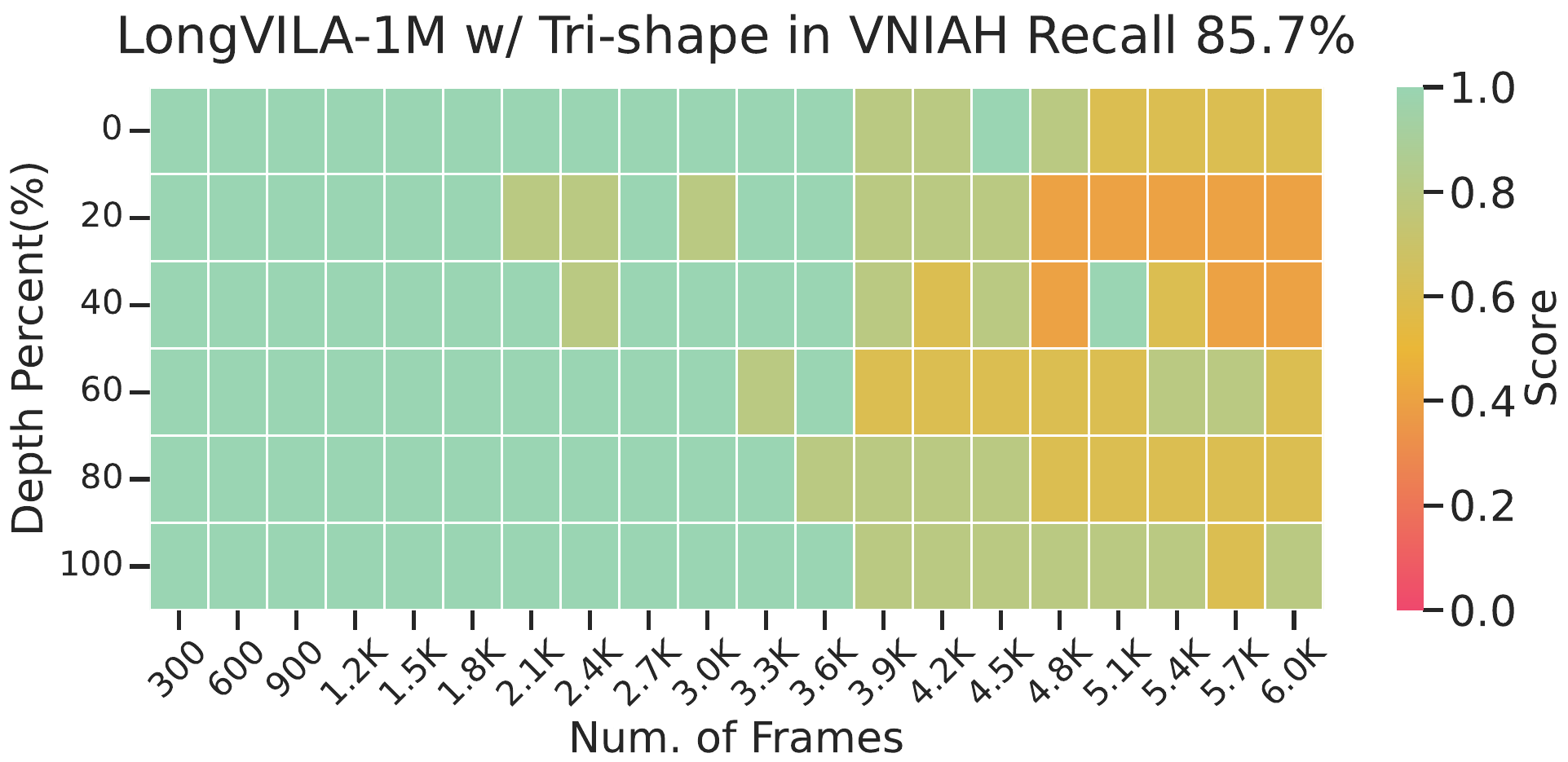}}\\
  \subfloat[SF-fixed]{
    \label{sfig:needle_sf_landmark_vniah}
    \includegraphics[width=0.5\columnwidth]{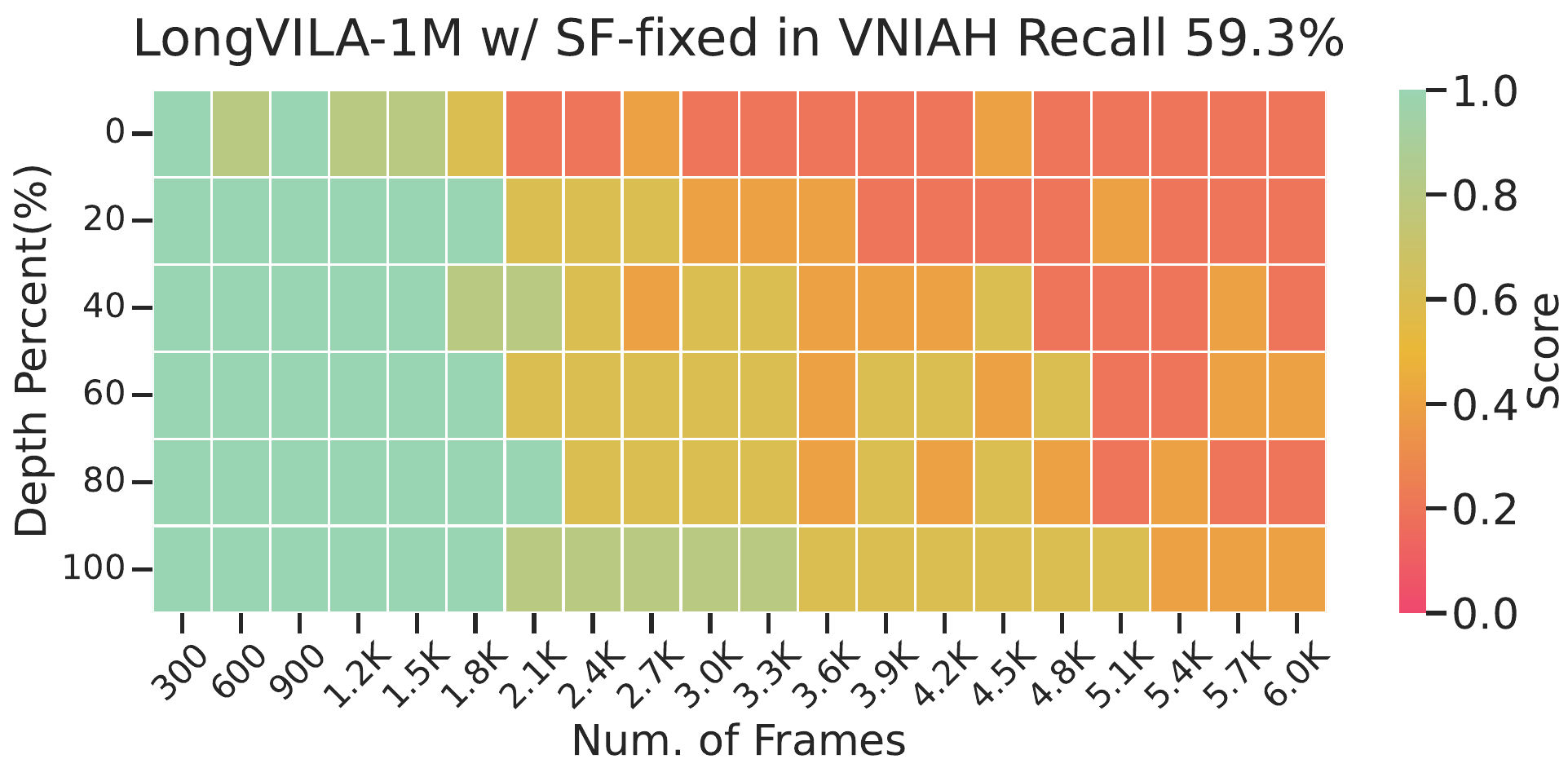}}
  \subfloat[SF-strided]{
    \label{sfig:needle_sf_strided_vniah}
    \includegraphics[width=0.5\columnwidth]{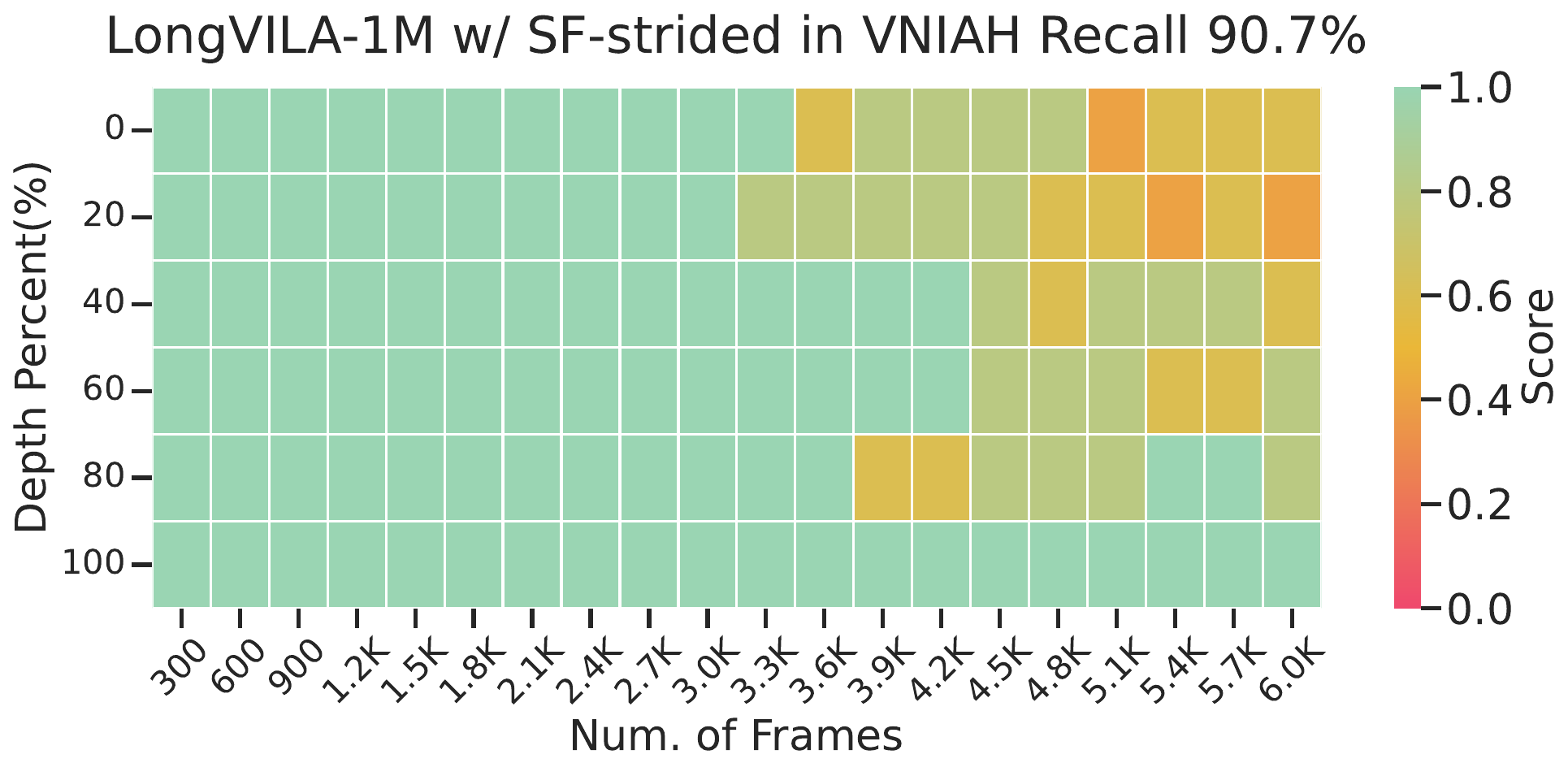}}\\
    \subfloat[MInference]{
    \label{sfig:needle_minference_vniah}
    \includegraphics[width=0.5\columnwidth]{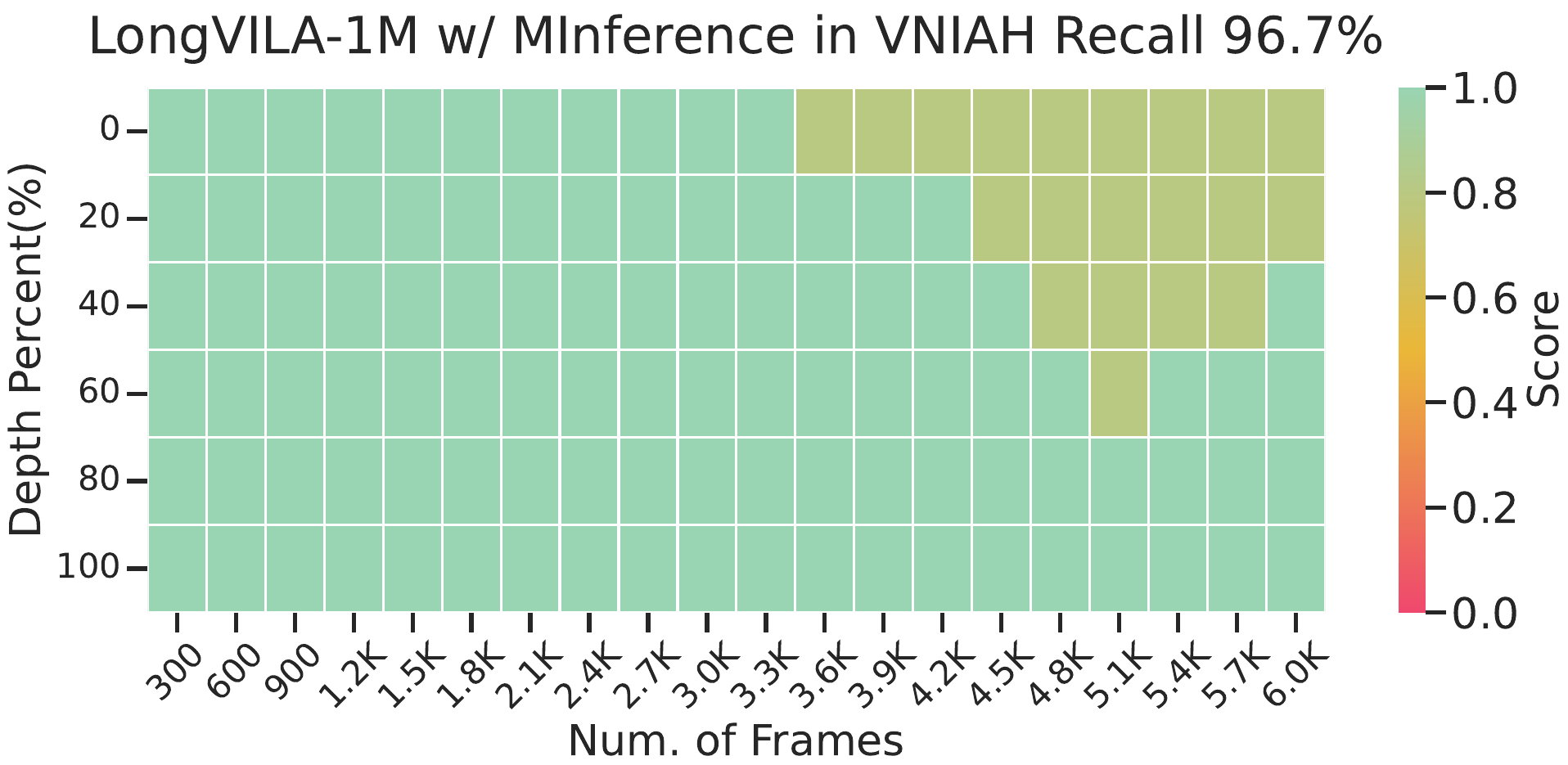}}\\
  \caption{Video Needle In A Haystack~\cite{zhang2024long} results using LongVila-Qwen2-7B-1M~\cite{xue2024longvila}.}
  \label{fig:needle_vniah}
\end{figure*}

\subsection{Additional Mixed-Modality Needle In A Haystack Results}

We further present the results of the Mixed-Modality Needle In A Haystack task with our baselines and the inter-modality variant of our method. The results of full atttenton and \methodall{} are shown in Fig.~\ref{fig:needle}.

\begin{figure*}[htb]
  \centering
  \subfloat[A-shape]{
    \label{sfig:needle_a_shape_MM-NIAH}
    \includegraphics[width=0.5\columnwidth]{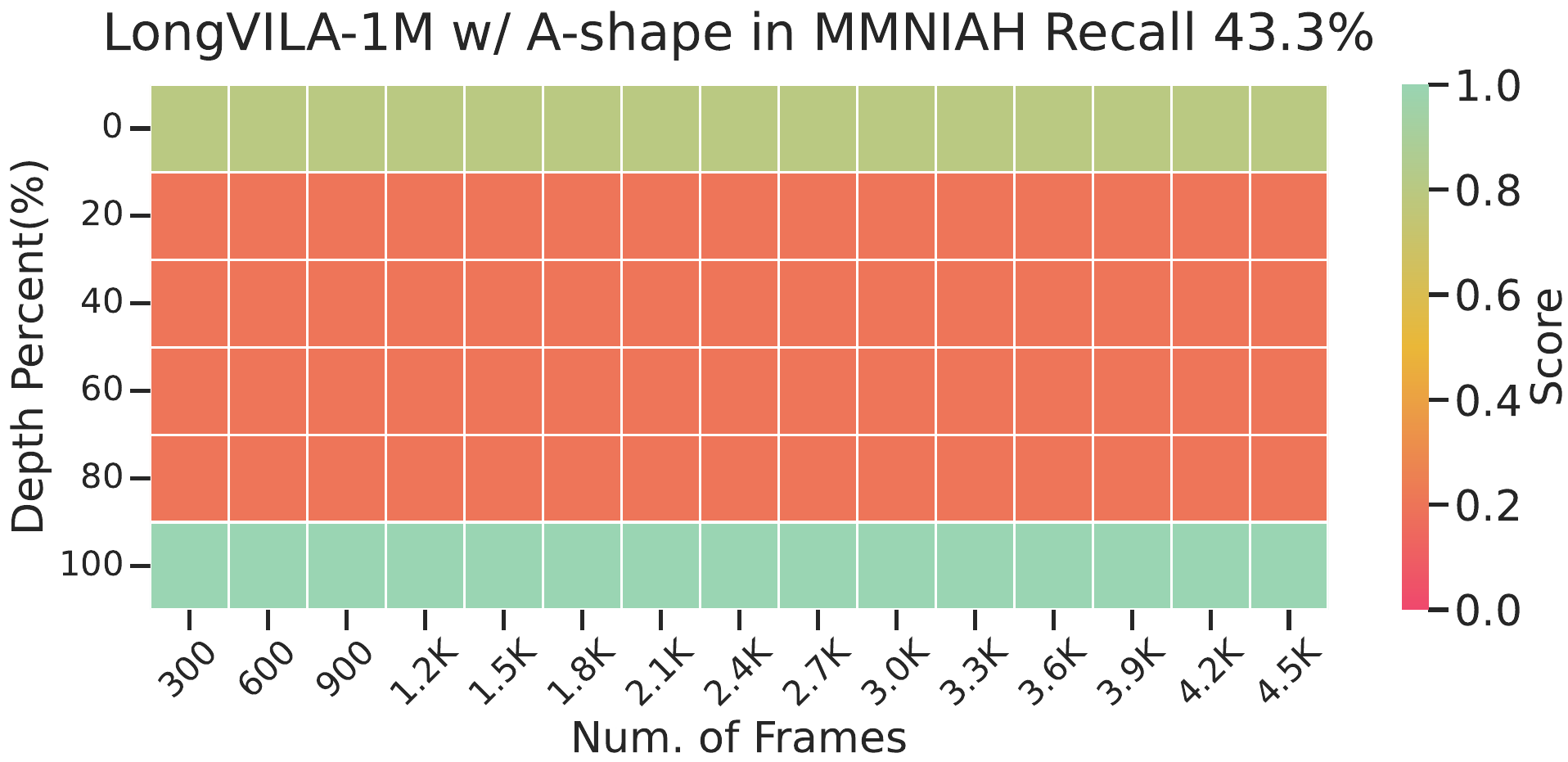}}
  \subfloat[Tri-shape]{
    \label{sfig:needle_tri_shape_MM-NIAH}
    \includegraphics[width=0.5\columnwidth]{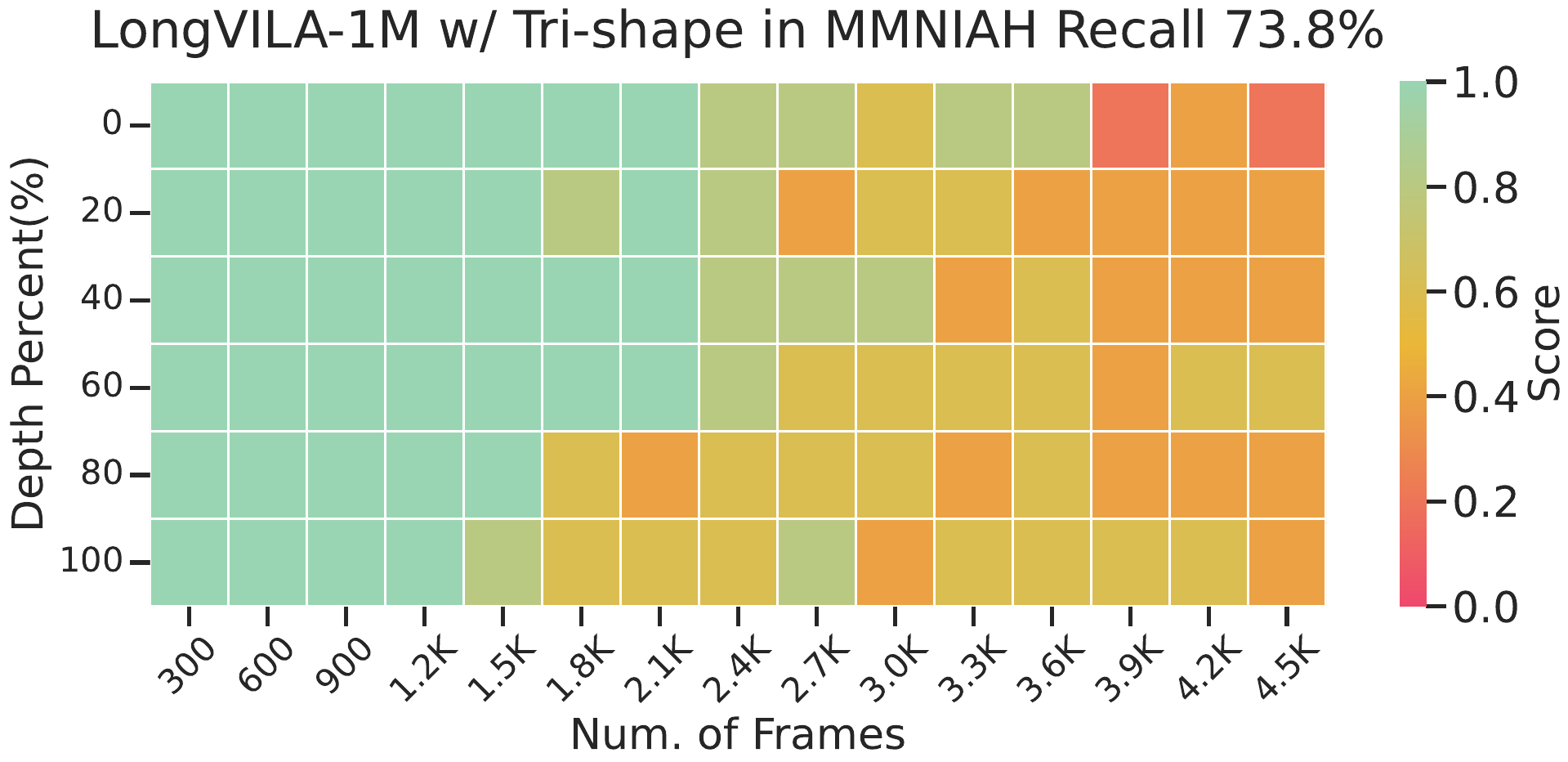}}\\
  \subfloat[MInference]{
    \label{sfig:needle_minference_MM-NIAH}
    \includegraphics[width=0.5\columnwidth]{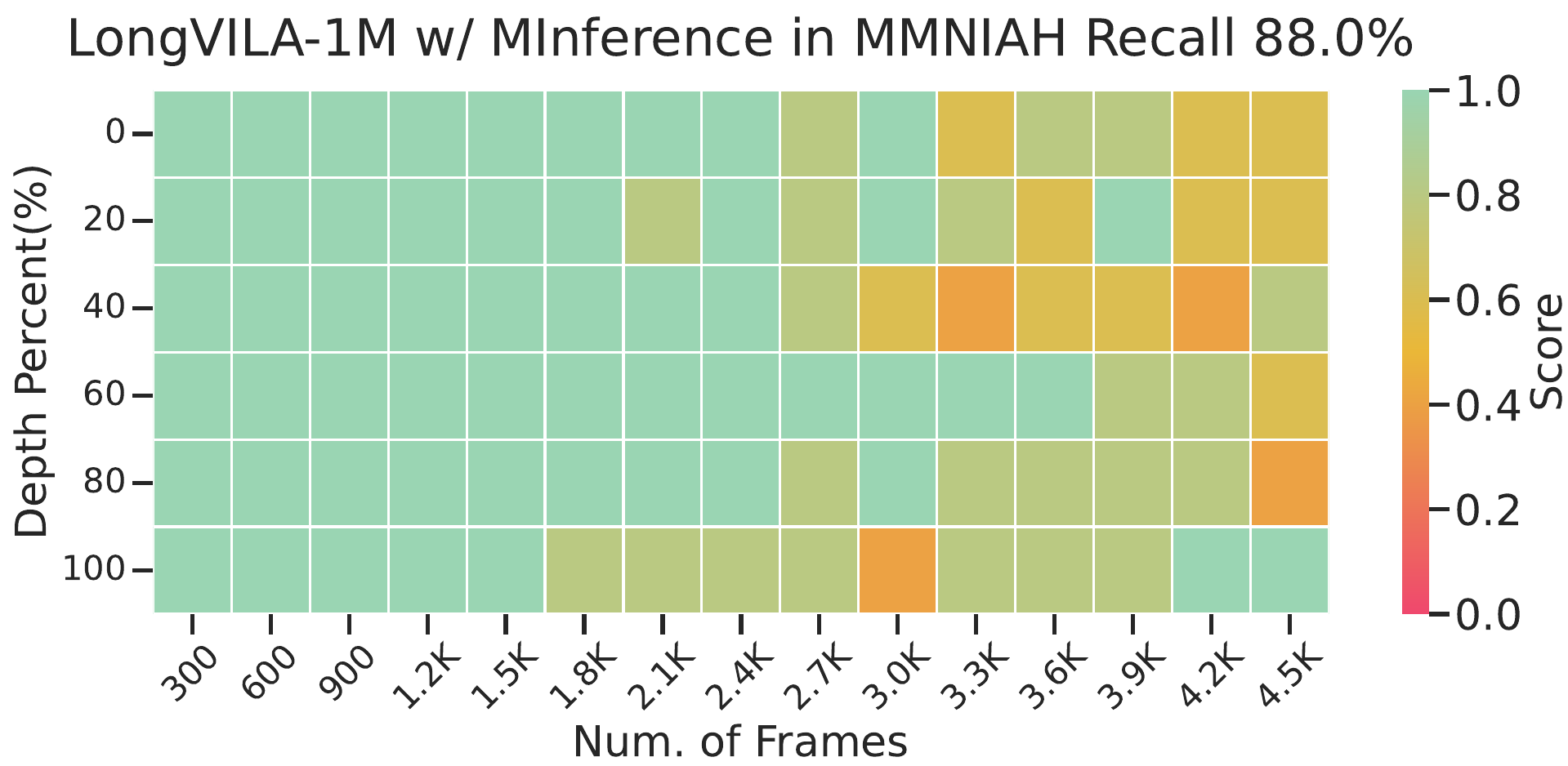}}
  \subfloat[\methodall{} w/o Inter-modality]{
    \label{sfig:needle_mminference_inter_MM-NIAH}
    \includegraphics[width=0.5\columnwidth]{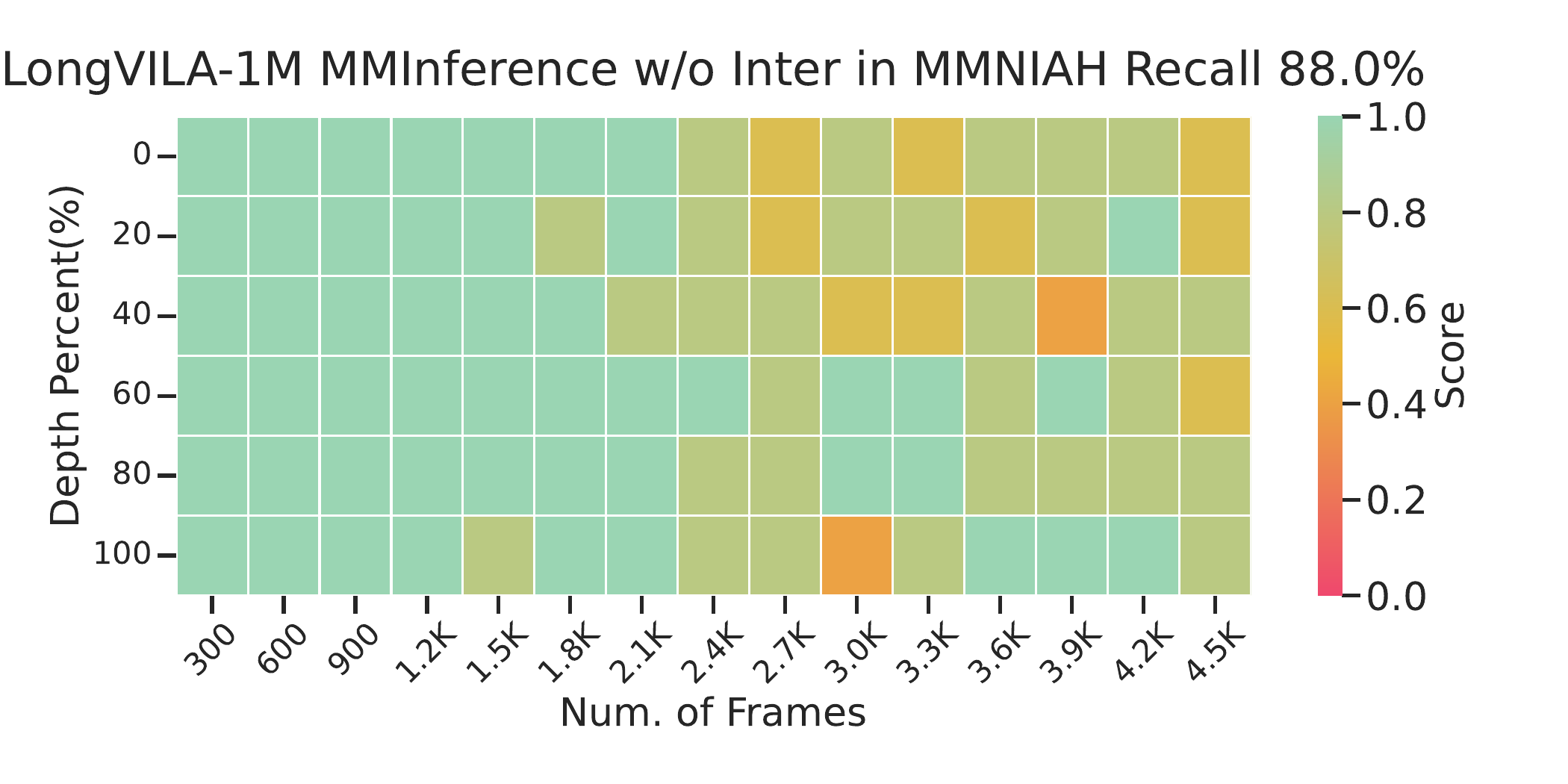}}\\
  \caption{Mixed-Modality Needle In A Haystack results using LongVila-Qwen2-7B-1M~\cite{xue2024longvila}.}
  \label{fig:needle_mmniah}
\end{figure*}

\subsection{Latency Breakdown}

As shown in Fig.~\ref{fig:latency_breakdown_detail}, we present the micro-benchmark results of various sparse attention methods across different context lengths.

\begin{figure*}[b!]
\vspace{-5pt}
  \centering
  \subfloat[Natten]{
    \label{sfig:natten}
    \includegraphics[width=0.3\columnwidth]{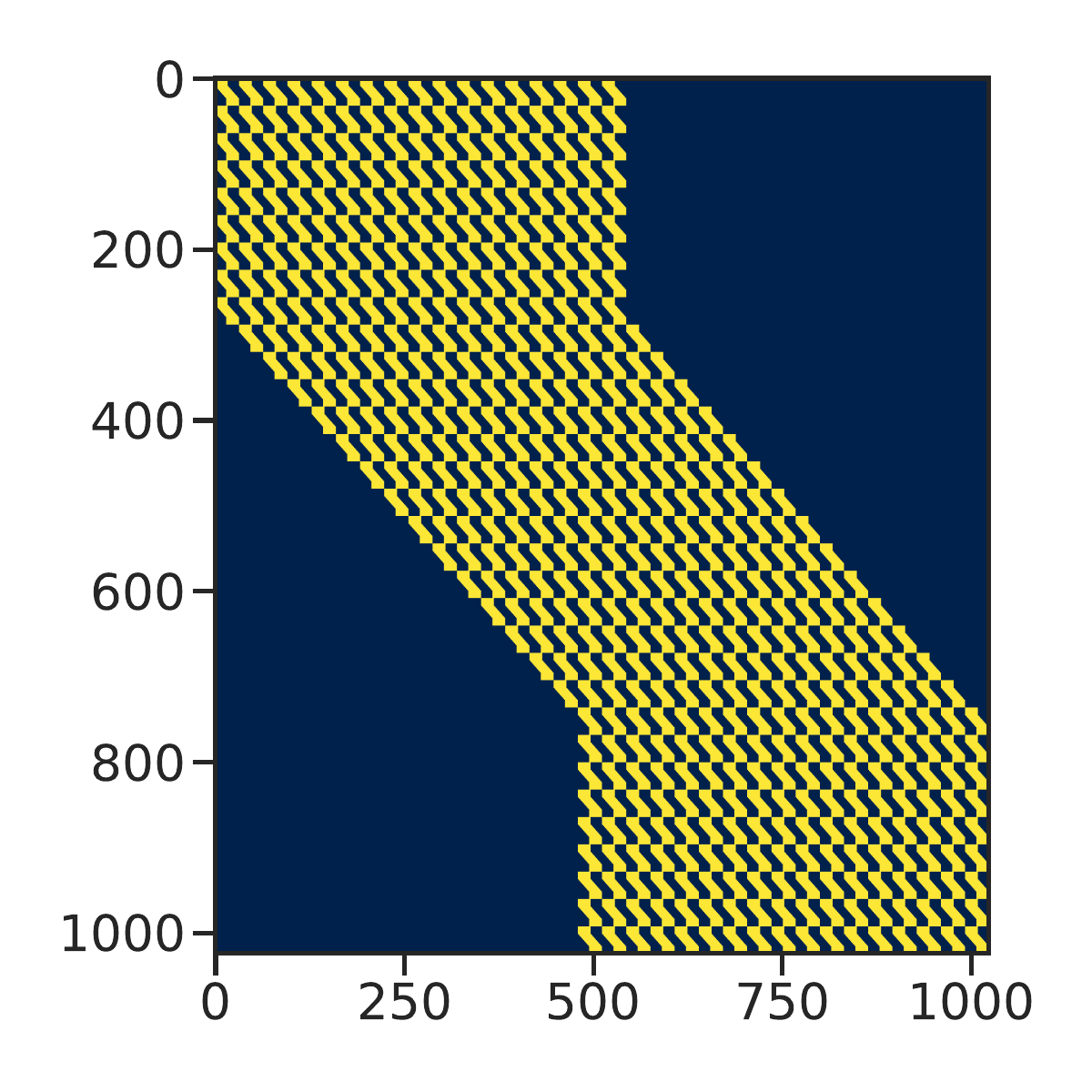}}
    \hspace{30pt}
  \subfloat[Permutated Natten]{
    \label{sfig:natten_permutated}
    \includegraphics[width=0.3\columnwidth]{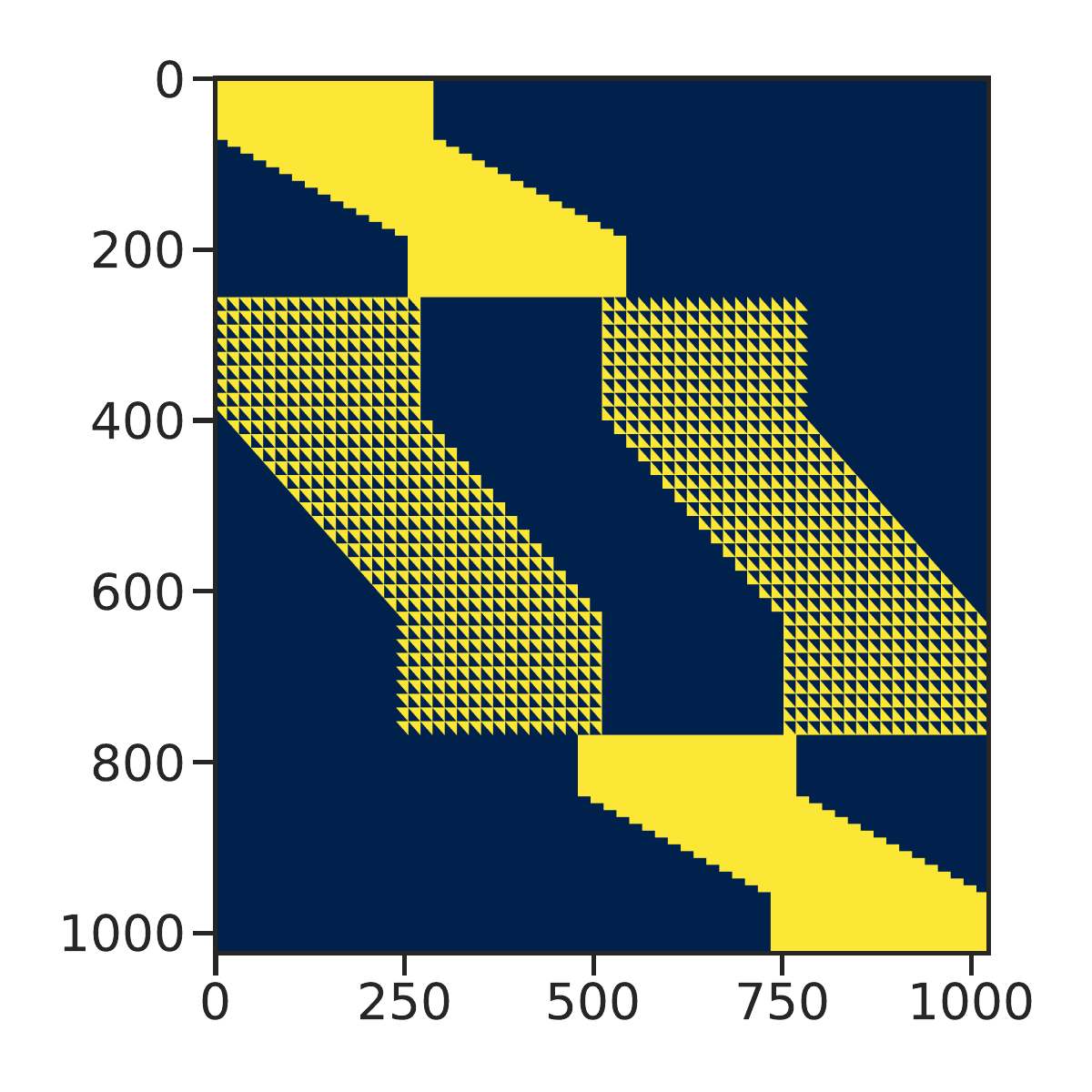}}\\
  \caption{Permutation-based implementation of 2D/3D sliding window attention~\cite{hassani2023neighborhood} enables efficient sparse attention optimization for DiT architectures.}
  \label{fig:dit}
\end{figure*}

\subsection{VS Pattern vs. Grid Pattern}

Both VS pattern and Grid pattern achieve strong performance on video understanding and V-NIAH tasks. However, due to the grid attention pattern observed in VLMs, the overlap between blocks covered by diagonal lines in the VS pattern is minimal, reducing sparsity within the kernel. This explains why VS pattern exhibits significantly higher latency compared to Grid pattern.
Additionally, leveraging permutation-based optimization effectively reduces the number of blocks involved in kernel computation, thereby lowering latency while maintaining comparable performance.

\section{Sparse Attention in DiT}

\begin{figure*}[htb]
  \centering
    \includegraphics[height=0.35\columnwidth]{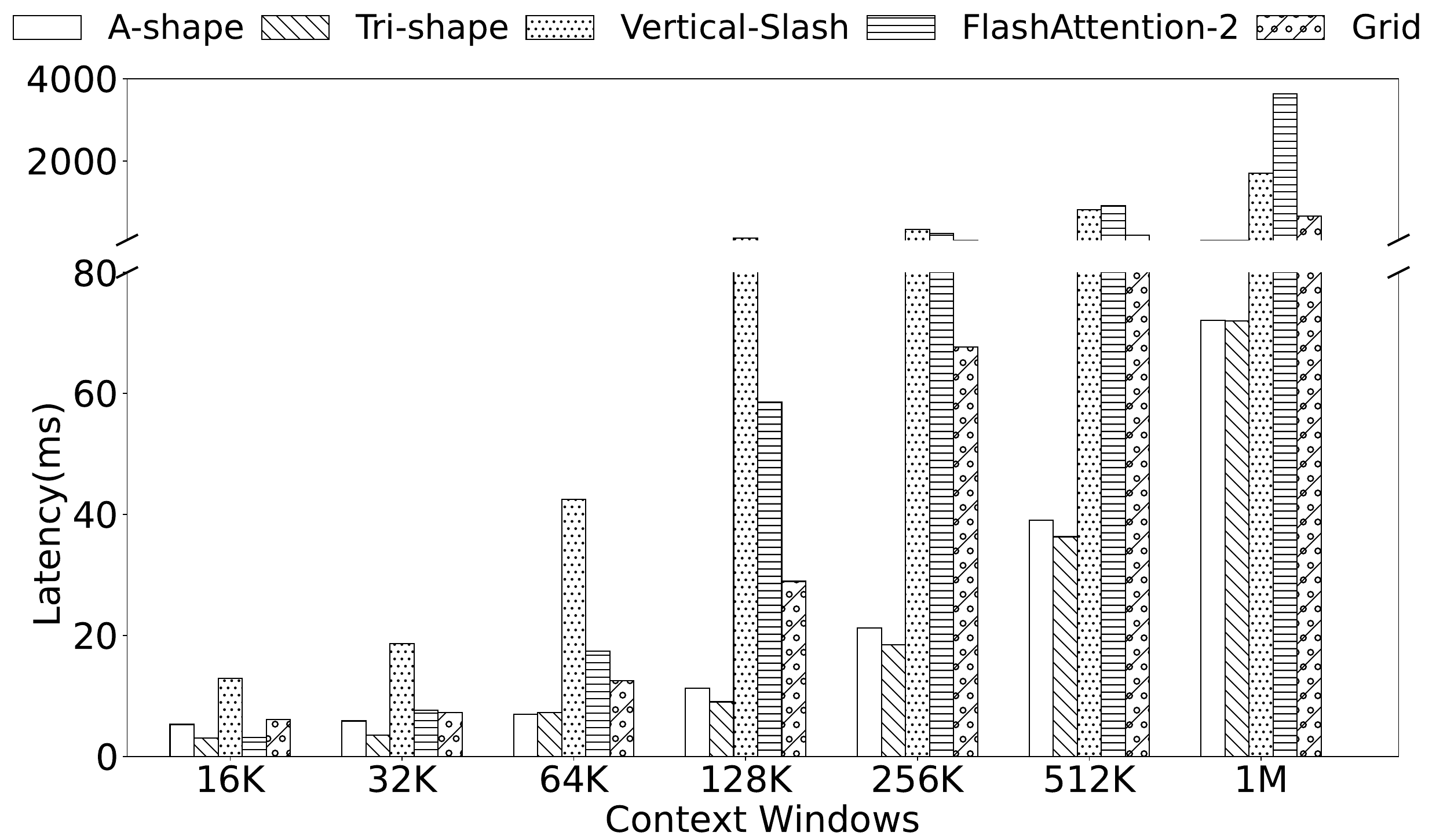}
  \caption{The latency breakdown of a single attention kernel for four sparse attention patterns and FlashAttention~\cite{dao2024flashattention2} across different context windows in a single A100, including the index time for dynamic sparse approximation and building dynamic sparsity. At 1M tokens, the latency for Grid is 358ms.}
  \label{fig:latency_breakdown_detail}
\end{figure*}

Recently, many efficient DiT methods~\cite{hassani2023neighborhood,xi2025sparse,zhang2025fast,xu2025xattention,ding2025efficient} have adopted sparse attention to accelerate long video generation. We note that these methods can also benefit from permutation-based transformations to achieve kernel-efficient implementations. For example, the 2D/3D sliding window attention in NATTEN can be converted into dense tensor core computation via permutation, as illustrated in Fig.~\ref{fig:dit}. Similarly, the temporal head in Sparse VideoGen~\cite{xi2025sparse} and the anti-diagonal structure in xAttention~\cite{xu2025xattention} can be restructured through permutation to enable sparse loading with dense computation, significantly speeding up DiT inference, especially in long-context scenarios.

\section{Kernel Implementation}

As shown in Algorithms~\ref{alg:grid_fa}, \ref{alg:q_boundary_fa}, and \ref{alg:2d_boundary_fa}, we provide implementation details of the FlashAttention-based kernels. The Grid-shape kernel in Algorithm~\ref{alg:grid_fa} integrates block-sparse FlashDecoding~\cite{flashdecoding}, which sparsifies the query loading, with block-sparse FlashAttention-2, which sparsifies the key loading. The Q-Boundary kernel in Algorithm~\ref{alg:q_boundary_fa} introduces sparsity along the query dimension using FlashAttention-2~\cite{dao2024flashattention2}, while the 2D-Boundary kernel in Algorithm~\ref{alg:2d_boundary_fa} applies sparsity along both the query and key dimensions.

\restylefloat{algorithm}

\begin{figure}[b!]
\centering
\begin{algorithm}[H]
\captionsetup[algorithm]{singlelinecheck=off}
\caption{Grid-Shape Flash Attention}
\label{alg:grid_fa}
\begin{minipage}[t]{0.48\textwidth}
\vspace{0pt}
\begin{algorithmic}
    \STATE {\bfseries Input:} 
    $\boldsymbol{Q}, \boldsymbol{K}, \boldsymbol{V} \in \mathbb{R}^{S \times d_h}$, 
    block size $B$, \diff{stride size $\sigma$, query start index $s_q$, key start index $s_k$}

    \STATE Scale $\tau \gets \sqrt{\frac1{d_h}}$
    \STATE Initialize $\boldsymbol{O} \gets (0)^{S \times d_h} \in \mathbb{R}^{S \times d_h}$

    \LineComment{Parallelized in GPU\\}
    \STATE{\textit{\diff{\# Sparse load in $\boldsymbol{Q}$ using FlashDecoding}}}
    \FOR{$i \gets 1$ to \diff{$N_\sigma$}}
        \STATE \diff{Load $\boldsymbol{Q}_{\text{chip}} 
        \gets \boldsymbol{Q}^{[i \times B:(i + 1)\times B]\times \sigma + s_q} \in \mathbb{R}^{B \times d_h}$}
        \STATE Initialize $\boldsymbol{O}_{\text{chip}} \gets (0)^{B \times d_h} \in \mathbb{R}^{B \times d_h}$
        \STATE Initialize $\boldsymbol{m} \gets (-\inf)^{B} \in \mathbb{R}^{B}$
        \STATE Initialize $\boldsymbol{l} \gets (0)^{B} \in \mathbb{R}^{B}$

        \LineComment{Loop in $\boldsymbol{K}$}
        \FOR{$j \gets 1$ to \diff{$M$}}
            \STATE \diff{Load $\boldsymbol{K}_{\text{chip}}
        \gets \boldsymbol{K}^{[j \times B:(j + 1)\times B]} \in \mathbb{R}^{B \times d_h}$}
        \STATE \diff{Load $\boldsymbol{V}_{\text{chip}}
        \gets \boldsymbol{V}^{[j \times B:(j + 1)\times B]} \in \mathbb{R}^{B \times d_h}$}
            \STATE $\boldsymbol{S} \gets \tau\boldsymbol{Q}_{\text{chip}}\boldsymbol{K}_{\text{chip}}^T$
            \STATE $\boldsymbol{S} \gets \mathrm{mask}(\boldsymbol{S})$
            \STATE $\boldsymbol{m}^i_{new} \gets \mathrm{max}(\boldsymbol{m}^i, \mathrm{rowmax}(\boldsymbol{S})) \in \mathbb{R}^{B}$
            \STATE $\boldsymbol{S} \gets \boldsymbol{S} - \boldsymbol{m}^i_{new}$
            \STATE $\boldsymbol{P} \gets \mathrm{exp}(\boldsymbol{S})$
            \STATE $\boldsymbol{l}^i_{new} \gets \mathrm{rowsum}(\boldsymbol{S}))$
            \STATE $\boldsymbol{\alpha} \gets \mathrm{exp}(\boldsymbol{m}^i - \boldsymbol{m}^i_{new})$
            \STATE $\boldsymbol{l}^i \gets \boldsymbol{\alpha}\boldsymbol{l}^i + \boldsymbol{l}^i_{new}$
            \STATE $\boldsymbol{O}_{\text{chip}} \gets \boldsymbol{\alpha}\boldsymbol{O}_{\text{chip}} + \boldsymbol{P}\boldsymbol{V}_{\text{chip}}$
        \ENDFOR

        \LineComment{Write outputs}
        \STATE $\boldsymbol{O}_{\text{chip}} \gets \mathrm{diag}(\boldsymbol{l}^i)^{-1}\boldsymbol{O}_{\text{chip}}$
        \STATE Save $\boldsymbol{O}_i \gets \boldsymbol{O}_{\text{chip}}$
    \ENDFOR
\end{algorithmic}
\end{minipage}
\hfill
\begin{minipage}[t]{0.48\textwidth}
\vspace{0pt}
\begin{algorithmic}
    \STATE{\textit{\diff{\# Sparse load in $\boldsymbol{K}$ using FlashAttention}}}
    \FOR{$i \gets 1$ to \diff{$N$}}
        \STATE \diff{Load $\boldsymbol{Q}_{\text{chip}} 
        \gets \boldsymbol{Q}^{[i \times B:(i + 1)\times B]} \in \mathbb{R}^{B \times d_h}$}
        \STATE Initialize $\boldsymbol{O}_{\text{chip}} \gets (0)^{B \times d_h} \in \mathbb{R}^{B \times d_h}$
        \STATE Initialize $\boldsymbol{m} \gets (-\inf)^{B} \in \mathbb{R}^{B}$
        \STATE Initialize $\boldsymbol{l} \gets (0)^{B} \in \mathbb{R}^{B}$

        \LineComment{Loop in $\boldsymbol{K}$}
        \FOR{$j \gets 1$ to \diff{$M_\sigma$}}
            \STATE \diff{Load $\boldsymbol{K}_{\text{chip}}
        \gets \boldsymbol{K}^{[j \times B:(j + 1)\times B]\times \sigma + \sigma \times s_k} \in \mathbb{R}^{B \times d_h}$}
            \STATE \diff{Load $\boldsymbol{V}_{\text{chip}}
        \gets \boldsymbol{V}^{[j \times B:(j + 1)\times B]\times \sigma + \sigma \times s_k} \in \mathbb{R}^{B \times d_h}$}
            \STATE $\boldsymbol{S} \gets \tau\boldsymbol{Q}_{\text{chip}}\boldsymbol{K}_{\text{chip}}^T$
            \STATE $\boldsymbol{S} \gets \mathrm{mask}(\boldsymbol{S})$
            \STATE $\boldsymbol{m}^i_{new} \gets \mathrm{max}(\boldsymbol{m}^i, \mathrm{rowmax}(\boldsymbol{S})) \in \mathbb{R}^{B}$
            \STATE $\boldsymbol{S} \gets \boldsymbol{S} - \boldsymbol{m}^i_{new}$
            \STATE $\boldsymbol{P} \gets \mathrm{exp}(\boldsymbol{S})$
            \STATE $\boldsymbol{l}^i_{new} \gets \mathrm{rowsum}(\boldsymbol{S}))$
            \STATE $\boldsymbol{\alpha} \gets \mathrm{exp}(\boldsymbol{m}^i - \boldsymbol{m}^i_{new})$
            \STATE $\boldsymbol{l}^i \gets \boldsymbol{\alpha}\boldsymbol{l}^i + \boldsymbol{l}^i_{new}$
            \STATE $\boldsymbol{O}_{\text{chip}} \gets \boldsymbol{\alpha}\boldsymbol{O}_{\text{chip}} + \boldsymbol{P}\boldsymbol{V}_{\text{chip}}$
        \ENDFOR

        \LineComment{Write outputs}
        \STATE $\boldsymbol{O}_{\text{chip}} \gets \mathrm{diag}(\boldsymbol{l}^i)^{-1}\boldsymbol{O}_{\text{chip}}$
        \STATE Save $\boldsymbol{O}_i \gets \boldsymbol{O}_{\text{chip}}$
    \ENDFOR
\end{algorithmic}
\end{minipage}

\end{algorithm}
\end{figure}

\begin{figure}[htb]

    \centering
    \begin{minipage}[t]{0.48\textwidth}
    \vspace{0pt}
    \centering
    \begin{algorithm}[H]
    \captionsetup[algorithm]{singlelinecheck=off}
    \caption{Q-Boundary Flash Attention}
    \label{alg:q_boundary_fa}
    \begin{algorithmic}
        \STATE {\bfseries Input:} 
        $\boldsymbol{Q}, \boldsymbol{K}, \boldsymbol{V} \in \mathbb{R}^{S \times d_h}$, 
        block size $B$, \diff{modality index $\boldsymbol{I}_m$, sparse attention kernel $\boldsymbol{\mathrm{Op}}_m$}
    
        \STATE Scale $\tau \gets \sqrt{\frac1{d_h}}$
        \STATE Initialize $\boldsymbol{O} \gets (0)^{S \times d_h} \in \mathbb{R}^{S \times d_h}$

        \LineComment{\diff{Loop modality} and parallelized in GPU}
        \FOR{\diff{$m \in \{\text{text}, \text{vision}, ...,\}$}}

    \FOR{{$i \gets 1$ to \diff{$N_m$}}}
        \STATE \diff{Load index $\boldsymbol{I}_{\text{chip}} 
        \gets \boldsymbol{I}_m^{[i \times B:(i + 1)\times B]} \in \mathbb{R}^{B}$}
        \STATE \diff{Load $\boldsymbol{Q}_{\text{chip}} 
        \gets \boldsymbol{Q}^{\boldsymbol{I}_{\text{chip}}} \in \mathbb{R}^{B \times d_h}$}
        \STATE Initialize $\boldsymbol{O}_{\text{chip}} \gets (0)^{B \times d_h} \in \mathbb{R}^{B \times d_h}$
        \STATE Initialize $\boldsymbol{m} \gets (-\inf)^{B} \in \mathbb{R}^{B}$
        \STATE Initialize $\boldsymbol{l} \gets (0)^{B} \in \mathbb{R}^{B}$

        \LineComment{Loop in $\boldsymbol{K}$ \diff{using modality sparse attention}}
        \STATE \diff{$\boldsymbol{O}_{\text{chip}}, \boldsymbol{m}, \boldsymbol{l} \gets \boldsymbol{\mathrm{Op}}_m(\boldsymbol{Q}_{\text{chip}}, \boldsymbol{K}, \boldsymbol{V}, \boldsymbol{O}_{\text{chip}}, \boldsymbol{m}, \boldsymbol{l})$}

        \LineComment{Write outputs \diff{w/ modality index}}
        \STATE $\boldsymbol{O}_{\text{chip}} \gets \mathrm{diag}(\boldsymbol{l}^i)^{-1}\boldsymbol{O}_{\text{chip}}$
        \STATE \diff{Save $\boldsymbol{O}_i^{\boldsymbol{I}_{\text{chip}}} \gets \boldsymbol{O}_{\text{chip}}$}
    \ENDFOR
    \ENDFOR
    \end{algorithmic}
    \end{algorithm}
    
    \end{minipage}
    \hfill
    \begin{minipage}[t]{0.5\textwidth}
    \vspace{0pt}
    
    \begin{algorithm}[H]
    \captionsetup[algorithm]{singlelinecheck=off}
    \caption{2D-Boundary Flash Attention}
    \label{alg:2d_boundary_fa}
    \begin{algorithmic}
        \STATE {\bfseries Input:} 
        $\boldsymbol{Q}, \boldsymbol{K}, \boldsymbol{V} \in \mathbb{R}^{S \times d_h}$, 
        block size $B$, \diff{modality index $\boldsymbol{I}_m$, sparse attention kernel $\boldsymbol{\mathrm{Op}}_m$}
    
        \STATE Scale $\tau \gets \sqrt{\frac1{d_h}}$
        \STATE Initialize $\boldsymbol{O} \gets (0)^{S \times d_h} \in \mathbb{R}^{S \times d_h}$

        \LineComment{\diff{Loop modality} and parallelized in GPU}
        \FOR{\diff{$m_q \in \{\text{text}, \text{vision}, ...,\}$}}

    \FOR{{$i \gets 1$ to \diff{$N_{m,q}$}}}
        \STATE \diff{Load index $\boldsymbol{I}_{\text{chip},q} 
        \gets \boldsymbol{I}_{m,q}^{[i \times B:(i + 1)\times B]} \in \mathbb{R}^{B}$}
        \STATE \diff{Load $\boldsymbol{Q}_{\text{chip}} 
        \gets \boldsymbol{Q}^{\boldsymbol{I}_{\text{chip},q}} \in \mathbb{R}^{B \times d_h}$}
        \STATE Initialize $\boldsymbol{O}_{\text{chip}} \gets (0)^{B \times d_h} \in \mathbb{R}^{B \times d_h}$
        \STATE Initialize $\boldsymbol{m} \gets (-\inf)^{B} \in \mathbb{R}^{B}$
        \STATE Initialize $\boldsymbol{l} \gets (0)^{B} \in \mathbb{R}^{B}$

        \LineComment{Loop in $\boldsymbol{K}$ \diff{and modality}}
        \FOR{\diff{$m_k \in \{\text{text}, \text{vision}, ...,\}$}}
        \FOR{$j \gets 1$ to \diff{$M_{m,k}$}}
        \STATE \diff{Load index $\boldsymbol{I}_{\text{chip},k} 
        \gets \boldsymbol{I}_{m,k}^{[j \times B:(j + 1)\times B]} \in \mathbb{R}^{B}$}
        \STATE \diff{Load $\boldsymbol{K}_{\text{chip}} 
        \gets \boldsymbol{K}^{\boldsymbol{I}_{\text{chip},k}} \in \mathbb{R}^{B \times d_h}$}
        \STATE \diff{Load $\boldsymbol{V}_{\text{chip}} 
        \gets \boldsymbol{V}^{\boldsymbol{I}_{\text{chip},k}} \in \mathbb{R}^{B \times d_h}$}

        \STATE \diff{$\boldsymbol{O}_{\text{chip}}, \boldsymbol{m}, \boldsymbol{l} \gets \boldsymbol{\mathrm{Op}}_m(\boldsymbol{Q}_{\text{chip}}, \boldsymbol{K}_{\text{chip}}, \boldsymbol{V}_{\text{chip}}, \boldsymbol{O}_{\text{chip}}, \boldsymbol{m}, \boldsymbol{l})$}
        \ENDFOR
        \ENDFOR

        \LineComment{Write outputs \diff{w/ modality index}}
        \STATE $\boldsymbol{O}_{\text{chip}} \gets \mathrm{diag}(\boldsymbol{l}^i)^{-1}\boldsymbol{O}_{\text{chip}}$
        \STATE \diff{Save $\boldsymbol{O}_i^{\boldsymbol{I}_{\text{chip},q}} \gets \boldsymbol{O}_{\text{chip}}$}
    \ENDFOR
    \ENDFOR
    \end{algorithmic}
    \end{algorithm}
    \end{minipage}
\end{figure}

\begin{figure*}[htb]
    \centering
    \subfloat[Qwen2.5-VL on EgoSchema]{
        \includegraphics[width=0.95\textwidth]{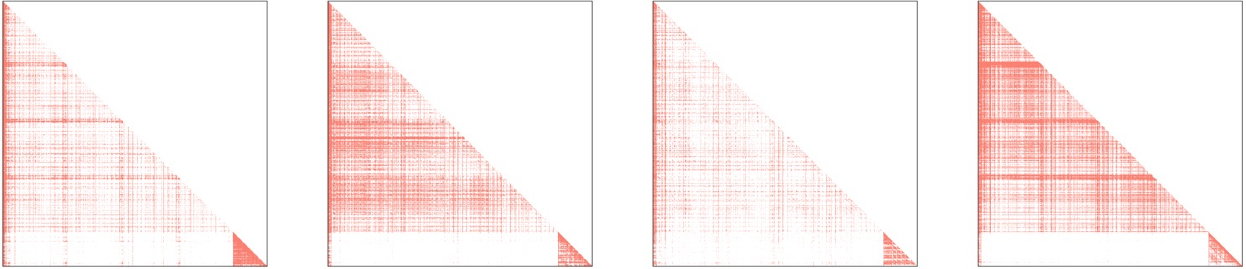}}\\
    \subfloat[VideoChat on EgoSchema]{
        \includegraphics[width=0.95\textwidth]{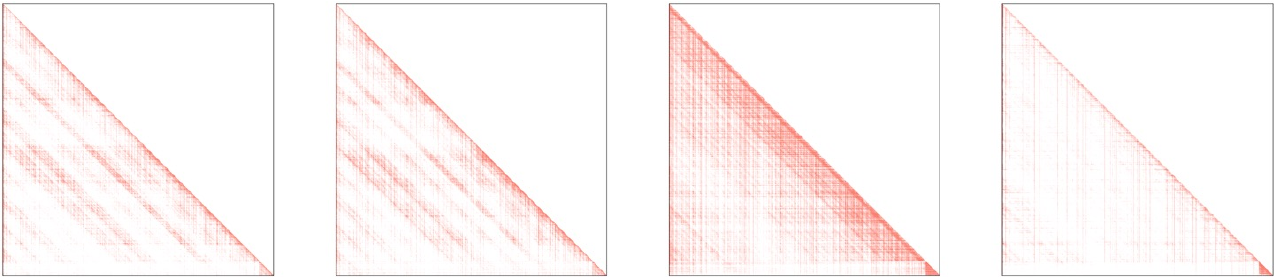}}
    \\
    \subfloat[Qwen2.5-VL on VideoMME]{
        \includegraphics[width=0.95\textwidth]{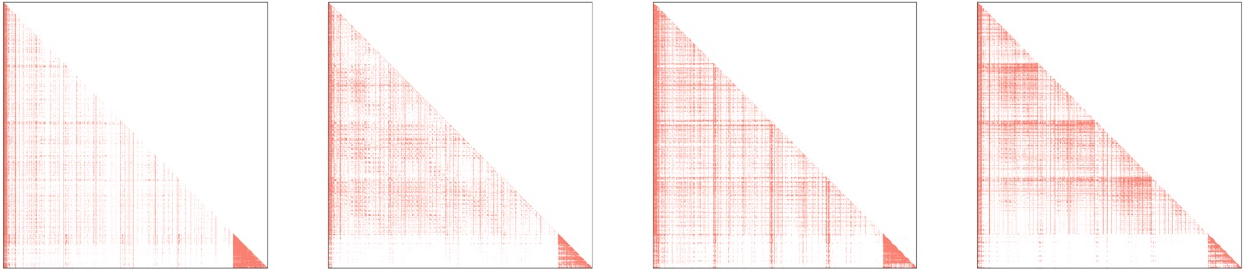}}\\
    \subfloat[VideoChat on VideoMME]{
        \includegraphics[width=0.95\textwidth]{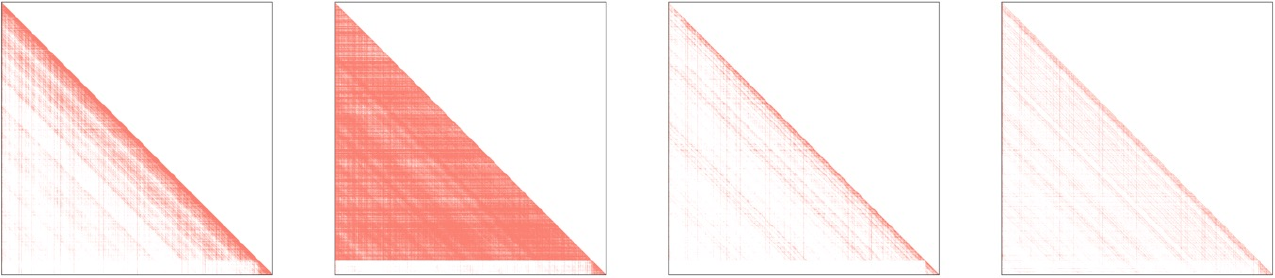}}
    \caption{Visualization of sparse attention patterns in Qwen2.5-VL with dynamic resolution input and VideoChat-Flash with visual token compression across different benchmarks.}
    \label{fig:additional_pattern_viz}
\end{figure*}

\begin{figure*}[htb]
    \centering
    \subfloat[Qwen2.5-VL on Mix-modality]{
        \includegraphics[width=0.95\textwidth]{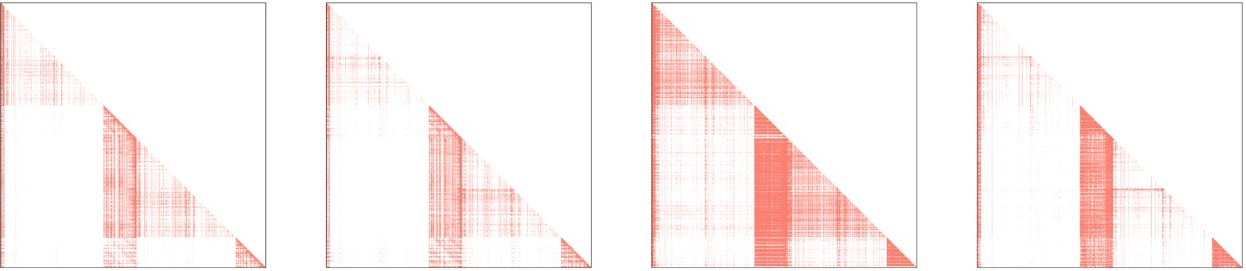}}\\
    \subfloat[VideoChat on Mix-modality]{
        \includegraphics[width=0.95\textwidth]{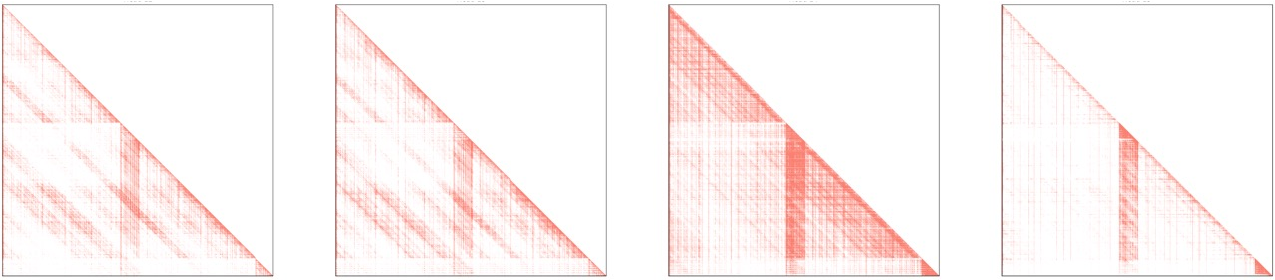}}
    \caption{Visualization of sparse attention patterns in Qwen2.5-VL with dynamic resolution input and VideoChat-Flash with visual token compression with mix-modality inputs.}
    \label{fig:additional_pattern_viz_mix}
\end{figure*}

\end{document}